%% file: main.tex
\documentclass[10pt,twocolumn,letterpaper]{article}

\usepackage{cvpr}
\usepackage[dvipsnames]{xcolor}
\input{preamble}

\definecolor{cvprblue}{rgb}{0.21,0.49,0.74}
\usepackage[pagebackref,breaklinks,colorlinks,allcolors=cvprblue]{hyperref}

\def\paperID{13204}
\def\confName{CVPR}
\def\confYear{2025}

\title{Spotting the Unexpected (STU): A 3D LiDAR Dataset for Anomaly Segmentation in Autonomous Driving}

\author{
Alexey Nekrasov$^1$
\and
Malcolm Burdorf$^1$
\and
Stewart Worrall$^2$
\and
Bastian Leibe$^1$
\and
Julie Stephany Berrio Perez$^2$ \\
\vspace{1em}
$^1$~RWTH Aachen University, $^2$~The University of Sydney\\
\vspace{0.75em}
{\url{https://vision.rwth-aachen.de/stu-dataset}}
}

\begin{document}
 \twocolumn[{%
 \renewcommand\twocolumn[1][]{#1}
 \maketitle
 \input{sec/teaser_fig}
 }]
 \input{sec/0_abstract}
 \input{sec/1_intro}
 \input{sec/2_rel_work}
 \input{sec/3_method}
 \input{sec/4_experiments}
 \input{sec/5_conclusion}

{
    \small
    \bibliographystyle{ieeenat_fullname}
    \bibliography{main}
}

\input{sec/X_suppl.tex}
\end{document}

%% file: preamble.tex
\usepackage{bm}
\usepackage{acro}

\usepackage{xspace}
\usepackage[accsupp]{axessibility}

\usepackage{array}
\usepackage{booktabs}
\usepackage{fontawesome}
\usepackage{xcolor,colortbl}
\definecolor{Gray}{gray}{0.9}
\definecolor{Textgray}{gray}{0.4}

\definecolor{notetext}{rgb}{0.7,0,0}
\definecolor{notetext}{rgb}{0.7,0,0}

\usepackage[nointegrals]{wasysym}
\usepackage{makecell}
\usepackage{balance}
\usepackage{multirow}
\usepackage{overpic}
\usepackage{bbm}
\usepackage{dsfont}
\usepackage[font=small,labelfont=bf]{caption}
\usepackage{color}
\usepackage{pifont}
\usepackage{etoolbox}
\usepackage{float}
\usepackage{tikz}
\usepackage{svg}
\usepackage{times}
\usepackage{epsfig}
\usepackage{graphicx}
\usepackage{amsmath}
\usepackage{amssymb}
\usepackage{comment}
\usepackage{soul}
\usepackage{listings}
\usepackage{lscape}
\usepackage{rotating}
\usepackage{enumitem}
\usepackage[utf8]{inputenc}
\usepackage{pgfplots}
\usepackage{import}
\usepackage{xifthen}
\usepackage{transparent}
\usepackage{tabularx}
\usepackage{tabu}
\usepackage{xcolor}
\usepackage{arydshln}
\DeclareUnicodeCharacter{2212}{−}
\usepgfplotslibrary{groupplots,dateplot}
\usetikzlibrary{patterns,shapes.arrows}
\usetikzlibrary{spy}
\pgfplotsset{compat=newest}

\usepackage{subcaption}
\usepackage{xfrac}
\usepackage{tcolorbox}

\usepackage{svg}
\newcommand{\orcid}[1]{\href{https://orcid.org/#1}{\includegraphics[width=10pt]{figures/orcid_logo}}}

\definecolor{m_red}{RGB}{255,209,209}
\definecolor{m_red_border}{RGB}{215,23,20}

\definecolor{m_orange}{RGB}{226,213,231}
\definecolor{m_orange_border}{RGB}{150,114,164}

\definecolor{m_blue}{RGB}{217,232,251}
\definecolor{m_blue_border}{RGB}{107,141,190}

\definecolor{m_yellow}{RGB}{255,242,205}
\definecolor{m_yellow_border}{RGB}{213,182,82}

\definecolor{m_gray}{RGB}{245,245,245}
\definecolor{m_gray_border}{RGB}{102,102,102}

\newcommand{\colorsquare}[1]{\protect\tikz{\path[draw=#1,fill=#1,thick,rounded corners=0.6pt] (0,0) rectangle (6pt,6pt);}}
\newcommand{\colorcirc}[1]{\protect\tikz{\fill[#1] (0,0) circle (2pt);}}

\lstdefinestyle{mypython}{
  language=python,
  breaklines=true,
  basicstyle=\fontsize{8}{12}\selectfont\ttfamily,
  keywordstyle=\bfseries\color{my_blue},
  linewidth=.99\textwidth,
}

\newcommand{\PAR}[1]{\vskip4pt \noindent {\bf #1~}}

\definecolor{darkgreen}{RGB}{0,255,0}
\definecolor{linkgreen}{RGB}{52,130,48}

\definecolor{Green}{RGB}{0, 180, 0}
\definecolor{Red}{RGB}{180, 0, 0}
\definecolor{Blue}{RGB}{30, 0, 180}
\newcommand{\cmark}{{\textcolor{Green}{\ding{51}}}}%
\newcommand{\xmark}{{\textcolor{Red}{\ding{55}}}}%

\newcolumntype{Y}{>{\centering\arraybackslash}X}
\newcolumntype{Z}{>{\raggedleft\arraybackslash}X}

\newif\ifmynotes
\mynotestrue
\definecolor{notetext}{rgb}{0.7,0,0}

\makeatletter
\def\adl@drawiv#1#2#3{%
        \hskip.5\tabcolsep
        \xleaders#3{#2.5\@tempdimb #1{1}#2.5\@tempdimb}%
                #2\z@ plus1fil minus1fil\relax
        \hskip.5\tabcolsep}
\newcommand{\cdashlinelr}[1]{%
  \noalign{\vskip\aboverulesep
           \global\let\@dashdrawstore\adl@draw
           \global\let\adl@draw\adl@drawiv}
  \cdashline{#1}
  \noalign{\global\let\adl@draw\@dashdrawstore
           \vskip\belowrulesep}}
\makeatother

%% file: sec/teaser_fig.tex
\vspace{-2em}
\begin{center}
    \captionsetup{type=figure}
    \includegraphics[trim={0cm 0cm 0cm 0.5cm},clip,width=\linewidth]{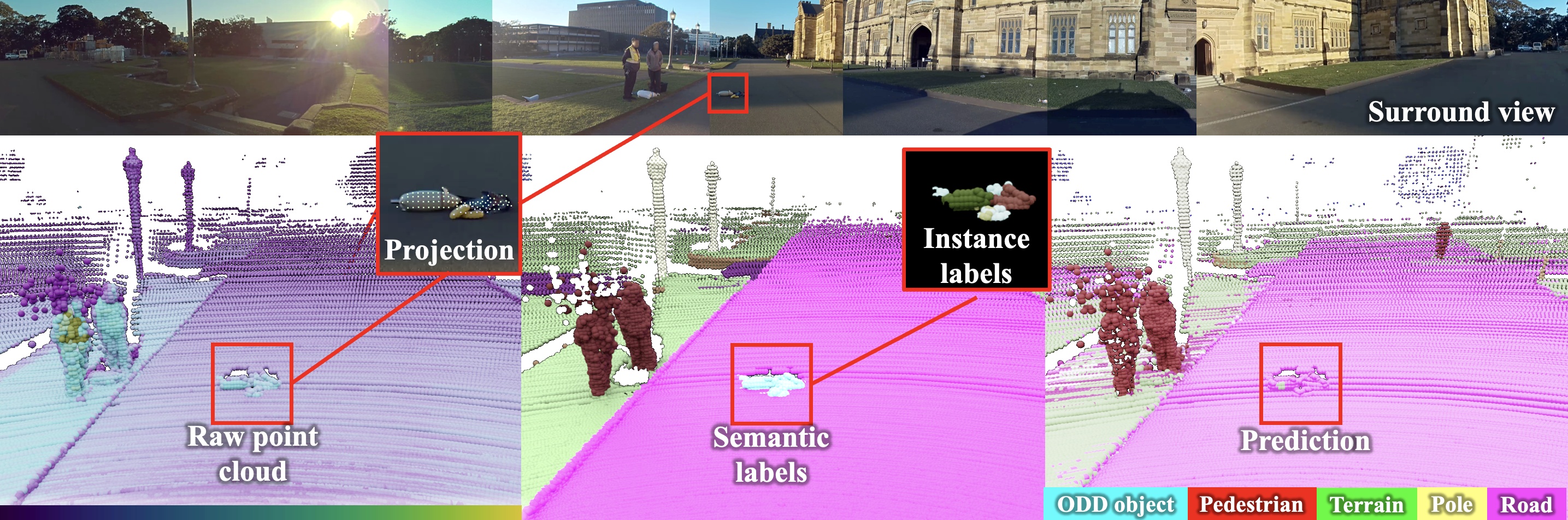}
    \captionof{figure}{
    \small 
    We present Spotting the Unexpected (STU) a novel anomaly segmentation dataset for autonomous driving.
    The dataset contains semantic and instance labels for out-of-distribution (OOD) objects, and includes surround-view setup with synchronized cameras.
    }
    \label{fig:teaser}
\end{center}

%% file: sec/0_abstract.tex
\begin{abstract}
    To operate safely, autonomous vehicles (AVs) need to detect and handle unexpected objects or anomalies on the road.
    While significant research exists for anomaly detection and segmentation in 2D, research progress in 3D is underexplored.
    Existing datasets lack high-quality multimodal data that are typically found in AVs. 
    This paper presents a novel dataset for anomaly segmentation in driving scenarios.
    To the best of our knowledge, it is the first publicly available dataset focused on road anomaly segmentation with dense 3D semantic labeling, incorporating both LiDAR and camera data, as well as sequential information to enable anomaly detection across various ranges.
    This capability is critical for the safe navigation of autonomous vehicles.
    We adapted and evaluated several baseline models for 3D segmentation, highlighting the challenges of 3D anomaly detection in driving environments. 
    Our dataset and evaluation code will be openly available, facilitating the testing and performance comparison of different approaches.
\end{abstract}

%% file: sec/1_intro.tex
\section{Introduction}
\label{sec:intro}

Autonomous vehicles have become an increasingly important part of modern transportation systems.
Despite significant advances, these vehicles continue to face challenges in accurately detecting unexpected objects on the road.
One particular challenge is road debris, which refers to objects that are not typically part of the roadway environment and pose significant safety risks.
According to the AAA Foundation for Traffic Safety~\cite{road_debri_report}, road debris was a contributing factor in more than $50,000$ crashes, $10,000$ injuries, and $125$ fatalities annually between $2011$ and $2014$.
These objects can vary widely in size and appearance, making them particularly challenging to detect.
Addressing this issue is critical for ensuring the safe and reliable operation of autonomous vehicles, as it directly impacts their ability to handle anomalies in real-world driving conditions.

State-of-the-art anomaly segmentation and anomaly detection methods have received substantial attention in recent years~\cite{nayal2023rba,liang2022gmmseg,du2021vos}, especially in the context of autonomous driving~\cite{gasperini2023u3hs,grcic2022densehybrid,tian2022pebal,chan2021entropy,jung2021sml,delic2024uno}.
One of the first benchmarks for evaluating anomaly segmentation is Lost and Found~\cite{pinggera2016laf}, whose data serves as the basis for many complex benchmarks~\cite{blum2021fishyscapes,chan2021segmentmeifyoucan}.
These benchmarks provide a wide range of possible road anomalies and obstacles, and recent research has focused on exploring the temporal~\cite{maag2022two} and multimodal~\cite{anovox,li2022coda} nature of anomaly segmentation.

While current autonomous driving datasets~\cite{behley2019semantickitti,caesar2020nuscenes,sun2020waymo} use a setup with surround view cameras, radar, and high-resolution LiDAR sensors, existing anomaly segmentation and detection benchmarks focus primarily on image data~\cite{bogdollsurvey} and mostly use Cityscapes~\cite{cordts2016cityscapes} as the in-distribution dataset.
Image-first benchmarks often lack the comprehensive sensor setup found in modern autonomous vehicles, and overlook the use of 3D modality for anomaly segmentation.
Recent research that has begun to investigate the use of LiDAR~\cite{singh2020lidar} sensor for anomaly segmentation and detection, lacks high-resolution sensors found in modern autonomous vehicles.
Several approaches have sought to find anomalies in validation sets of existing datasets~\cite{li2022coda,piroli2023lsvos} or have created datasets using simulation~\cite{anovox} due to the difficulty of collecting and annotating such data in the real world.
Despite these efforts, extending existing data or simulating anomalies only partially addresses the lack of accessible and high-quality data, shown in Table~\ref{tab:dataset_comparison}.

To address this issue, we analyzed the labels of two widely used in-distribution (ID) datasets~\cite{behley2019semantickitti, caesar2020nuscenes} to identify an appropriate set of anomalies that would not overlap with objects in the ID training or validation datasets.
This informed our creation of a real-world dataset containing challenging anomaly objects placed on roads, with annotated sequences in 3D LiDAR point clouds.
Our dataset consists of multiple high-resolution LiDAR sequences acquired in challenging environments, with point-level labels for individual anomaly objects (shown in Figure~\ref{fig:teaser}).
We acquired data in both rural and urban environments, with objects of varying degrees of difficulty, and include a number of real-world anomalies found on public roads.
In addition, to evaluate in-distribution performance, we annotated additional sequences without anomalies with inlier labels to train and evaluate models to ensure a small domain gap between our and SemanticKITTI~\cite{behley2019semantickitti} datasets.

To establish simple baselines for 3D LiDAR segmentation, we adapted the methods used in 2D anomaly segmentation~\cite{nayal2023rba,srivastava2014mcdropout,lakshminarayanan2017deepensemble,blum2021fishyscapes,hendrycks2018baseline} and applied them to the 3D domain, investigating anomaly segmentation methods for 3D LiDAR point clouds.
Our models, trained jointly on the SemanticKITTI dataset and our own dataset, show strong segmentation performance for in-distribution data and novel additional sequences.
However, we observed that the baseline methods struggled with out-of-distribution objects, and models tend to predict anomaly objects as inliers with a high degree of confidence.
We believe that our dataset will benefit the community by allowing the evaluation of anomaly segmentation methods in the LiDAR domain and further exploration of multimodal anomaly segmentation methods.

Our contributions provide a basis for the development of more advanced anomaly segmentation methods in the 3D LiDAR domain.
Specifically, we propose a new anomaly segmentation dataset, and perform an analysis of the recorded and annotated data.
The purpose of the paper is to provide practitioners with the necessary tools to evaluate their anomaly segmentation models.
We also provide a set of baselines that can serve as a reference for future challenges and research directions in this area.

%% file: sec/2_rel_work.tex
\section{Related work}

\subsection{Anomaly Datasets}
Several multimodal datasets for autonomous driving are available, including Waymo~\cite{sun2020waymo}, nuScenes~\cite{caesar2020nuscenes}, SemanticKITTI~\cite{behley2019semantickitti}, among others.
These datasets were collected in urban environments under typical driving conditions.
Although these datasets are relevant for evaluating various tasks related to autonomous vehicle operation, they lack data for uncommon scenarios.
One such scenario involves the presence of foreign objects on the road.
Although this situation is rare during typical driving, autonomous vehicles must be able to recognize and respond to it safely. 
Here, assigning a unique label to each distinct object is impractical, since datasets have a limited set of labels, and objects on the road can vary widely.
These objects are often grouped under a general category like ``other object'' and are excluded from the benchmarks due to their variability.

\input{tables/1_comparison}

In recent years, several datasets have been introduced to assess the detection of foreign objects on the road~\cite{blum2021fishyscapes,pinggera2016laf,chan2021segmentmeifyoucan,hendrycks2022streethazards,li2022coda,maag2022two,xu2024raod,nekrasov2024oodis}.
Some of these datasets have been developed using data augmentation techniques or simulations, but only a few are based on data collected with real-world equipment in real driving scenarios.
Simulations and augmentations are great tools, but they lack realism in terms of the sensor model~\cite{blum2021fishyscapes} and the sizes of the objects~\cite{hendrycks2022streethazards}.
However, the aforementioned datasets are limited to image data only.
Autonomous vehicles use LiDAR sensors because they offer greater range, robustness to lighting conditions, and more precise 3D information about the environment compared to monocular depth estimations or stereo cameras.
Therefore, incorporating 3D information is essential for accurately assessing road anomalies.

Three datasets that integrate camera and LiDAR data are SOD~\cite{gupta2018mergenet}, TOR4D~\cite{wong2019osis} and CODA~\cite{li2022coda}.
The CODA series is derived from validation sets of existing autonomous vehicle datasets.
Still, it has the limitation that anomalous objects (\eg, traffic cones, barriers) may also appear in the training set, which is acceptable for the task of open-set segmentation, but contradicts the task of anomaly segmentation.
Recently, a few methods for 3D anomaly detection have been proposed~\cite{piroli2023lsvos,kosel2024mmod3d,mood3D}. However, these methods use validation sequences to evaluate anomaly detection capabilities.
The SOD dataset offers depth information for small road obstacles using a low-resolution $16$-beam LiDAR, and that introduces a large domain gap between the training and test data.
TOR4D and Rare4D \cite{wong2019osis} are the only datasets that contain real-world LiDAR data to benchmark road anomalies; however, these datasets are proprietary and not accessible to the public.
We release out-of-distribution validation and in-distribution training sequences with a permissive license and open the private test set to submissions from the public.

\subsection{Open-Set Segmentation in 3D}
Open-set and unsupervised segmentation are closely related to the out-of-distribution detection task.
Multiple methods were proposed for unsupervised segmentation~\cite{nunes202223duis,semoli} and open-set prediction in 3D~\cite{lps,wong2019osis}.
Recent evaluations in 2D anomaly segmentation~\cite{gasperini2023u3hs} have demonstrated that methods such as OSIS~\cite{wong2019osis} and EOPSN~\cite{hwang2021eopsn} perform poorly on anomaly objects.
This weakness stems from open-set methods' reliance on the presence of unknown instances inside the void regions during training, which may not generalize well to real-world scenarios.
Similarly, MLUC~\cite{cen2021openseta} assesses open-set performance on the UDI~\cite{cen2021openseta} dataset, but this dataset is also not publicly accessible and makes the same assumptions as OSIS methods.
In addition, open-set methods typically remove the ground plane~\cite{nunes202223duis} to obtain clusters of potential objects, but ground plane removal~\cite{patchwork} methods are often sensitive to chosen hyperparameters and yield noisy results, missing many smaller objects.
For the evaluation of anomaly segmentation, it is important to have a test dataset that is free of anomaly objects which might occur during training.
In contrast to open-set methods, our work focuses on collecting and densely annotating real-world anomaly objects, ensuring that there is no intersection between in-distribution training sets and out-of-distribution test sets.

\subsection{Anomaly Segmentation in Autonomous Driving}
To tackle the problem of anomaly segmentation, multiple works~\cite{blum2021fishyscapes,bogdollsurvey,chan2021segmentmeifyoucan} have recently been proposed for the problem of 2D anomaly segmentation.
Earlier works~\cite{hendrycks2018baseline,chan2021entropy,lakshminarayanan2017deepensemble,tian2022pebal,grcic2022densehybrid,liang2022gmmseg,blum2021fishyscapes} were applied to per-pixel segmentation models such as DeepLab~\cite{chen2018deeplabv3}; however, recently many works were focused on MaskFormer-style models~\cite{cheng2021maskformer,cheng2021mask2former}. 
A line of work~\cite{delic2024uno,nayal2023rba,rai2023mask2anomaly} proposed different methods to adapt Mask2Former, exploring the mechanism of mask prediction, where multiple masks might ignore certain regions of an image.
These models provide more reliable anomaly scoring mechanisms not limited to per-pixel predictions and achieve higher performance on the anomaly segmentation task.
For our benchmark, we decided to evaluate methods that also use MaskFormer-style models and investigate the application of the Mask4Former~\cite{yilmaz2024mask4former} model for 3D anomaly segmentation.
We obtain all our scores using a MaskFormer-Style model, laying the foundation for future baselines.

%% file: tables/1_comparison.tex
\begin{table*}[t]
\centering
\setlength{\tabcolsep}{2pt}
\caption{\small Comparison of our dataset with other community datasets~\cite{bogdollsurvey}. SM = Semantic Mask, IM=Instance Mask, BB=Bounding Box}
\label{tab:dataset_comparison}

\scriptsize
\begin{tabularx}{\textwidth}{l YY c YYYY c c}
 \toprule
 \textbf{Dataset} &
 \textbf{Camera} & 
 \textbf{\begin{tabular}[c]{@{}c@{}}LiDAR\\ Beams\end{tabular}} &
 \textbf{Source} &
 \textbf{\#ODD} &
 \textbf{Label} &
 \textbf{\begin{tabular}[c]{@{}c@{}}Size\\ Test/Val\end{tabular}} &
 \textbf{Temporal} &
 \textbf{\begin{tabular}[c]{@{}c@{}}Environments\\ Test Sequences\end{tabular}} &
 \textbf{Related Dataset} \\
 
\midrule
\textbf{Fishyscapes}~\cite{blum2021fishyscapes,fishyscapes_2} &&&& \multicolumn{1}{l}{} && \multicolumn{1}{l}{} &&& \\

FS Lost and Found
&
1 RBG &
\xmark &
\textcolor{Green}{Staged} &
1 &
SM &
275/100 &
\xmark &
n/a &
Lost and Found~\cite{pinggera2016laf} \\
 
FS Static &
1 RGB &
\xmark &
\textcolor{Orange}{Augmentated} &
1 &
SM &
1000/30 &
\xmark &
n/a &
CityScapes~\cite{cordts2016cityscapes} \\

\hline
\textbf{CAOS}~\cite{hendrycks2022streethazards} &&&& \multicolumn{1}{l}{} && \multicolumn{1}{l}{} & & & \\

StreetHazards &
1 RGB &
\xmark &
\textcolor{Red}{Simulated} &
1 &
SM &
1500 &
\cmark &
2 sim towns - 150 seq &
\\

BDD-Anomaly &
1 RGB &
\xmark &
\textcolor{Orange}{Validation Set} &
3 &
SM &
810 &
\xmark &
n/a &
BDD100K~\cite{yu2020bdd100k} \\
 
\hline
\textbf{SegmentMeIfYouCan}~\cite{chan2021segmentmeifyoucan} & & & & \multicolumn{1}{l}{} & & \multicolumn{1}{l}{} & & & \\

RoadAnomaly21 &
1 RGB &
\xmark &
\textcolor{Red}{Web Sourced} &
1 &
SM &
100/10 &
\xmark &
n/a &
\\

RoadObstacle21 &
1 RGB &
\xmark &
\textcolor{Green}{Staged} &
1 &
SM &
382/30 &
\cmark &
9 streets - 90 seq &
\\

\hline
\textbf{CODA}~\cite{li2022coda} & & & & \multicolumn{1}{l}{} & & \multicolumn{1}{l}{} & & & \\

CODA-KITTI &
1 RGB &
\cmark - 64 &
\textcolor{Orange}{Validation Set} &
6 &
BB &
309 &
\xmark &
n/a &
KITTI~\cite{geiger2013kitti} \\

CODA-nuScenes &
1 RGB &
\cmark - 32 &
\textcolor{Orange}{Validation Set} &
17 &
BB &
134 &
\xmark &
n/a & NuScenes \cite{caesar2020nuscenes} \\
 
CODA-ONCE &
1 RGB &
\cmark - 40 &
\textcolor{Orange}{Validation Set} &
32 &
BB &
1057 &
\xmark &
n/a &
ONCE~\cite{mao2021once} \\
 
CODA22-ONCE &
1 RGB &
\cmark - 40 &
\textcolor{Orange}{Validation Set} &
29 &
BB &
717 &
\xmark &
n/a &
ONCE~\cite{mao2021once} \\
 
CODA22-SODA10M &
1 RGB &
\xmark &
\textcolor{Orange}{Validation Set} &
29 &
BB &
\ul{4167} &
\xmark &
n/a &
SODA~\cite{han2021soda} \\
 
\hline
\textbf{Wuppertal ODD}~\cite{maag2022two} & & & & \multicolumn{1}{l}{} & & \multicolumn{1}{l}{} & & & \\

SOS &
1 RGB &
\xmark - dist &
\textcolor{Green}{Staged} &
13 &
IM &
1129 &
\cmark &
9 streets - 20 seq &
\\
 
CWL &
1 RGB &
\xmark - dist &
\textcolor{Red}{Simulated} &
18 &
IM &
1210 &
\cmark &
26 streets - 26 seq &
\\
 
\hline
\textbf{Lost and Found}~\cite{pinggera2016laf} &
1 Stereo &
\xmark &
\textcolor{Green}{Staged} &
42 &
SM &
1068 &
\cmark &
5 streets - 5 seq &
\\

\textbf{SOD}~\cite{singh2020lidar} &
1 RGB &
\cmark - 16 &
\textcolor{Green}{Staged} &
1 &
SM &
460/530 &
\xmark &
2 streets - 5 seq &
\\

\textbf{WD-Pascal}~\cite{bevandic2019simultaneous} &
1 RGB &
\xmark &
\textcolor{Red}{Simulated} &
1 &
SM &
70 &
\xmark &
n/a &
\\

\textbf{Vistas-NP}~\cite{vistas_np} &
1 RGB &
\xmark &
\textcolor{Orange}{Class exclusion} &
4 &
SM &
11167 &
\xmark &
n/a &
Mapillary Vistas \cite{neuhold2017mapillary} \\

\hline
\rowcolor{blue!10}
\textbf{STU (Ours)} &
\textbf{8 RGB} &
\textbf{\cmark - 128} &
\textbf{\textcolor{Green}{Staged} } &
\textbf{1} &
\textbf{IM} &
\multicolumn{1}{c}{\textbf{8022/1960}} &
\textbf{\cmark } &
\textbf{6 streets - 51 seq} &
SemKITTI~\cite{behley2019semantickitti}, P-CUDAL~\cite{tseng2025panopticcudal} \\ %
 
\bottomrule 
\end{tabularx}
\end{table*}

%% file: sec/3_method.tex
\section{The Spotting the Unexpected (STU) Dataset}
Our data collection platform consists of a rigid frame mounted on a vehicle and equipped with eight hardware-triggered cameras and a LiDAR sensor to ensure synchronized data collection.
The LiDAR and the cameras were calibrated using the method described in \cite{surabhi2019calibration}, with the calibration process repeated for each camera.
This setup follows the configuration discussed in Panoptic-CUDAL~\cite{tseng2025panopticcudal}.
We refer to the Supplementary Material for detailed information on the vehicle setup.

\subsection{Data collection.}
The data was collected using two sets of conditions: one in a naturalistic environment where objects were found on public roads, and one in a controlled environment.
In the naturalistic scenario, several hours of driving data were recorded, with specific events marked when objects appeared on the road.
Figure \ref{fig:data_naturalistic} shows a bucket on the road recorded while driving on a highway. 

\input{figures/3_data_collection/figure}
\input{figures/2_objects/figure}

The controlled environment data collection protocol was rigorous for safety reasons.
We employed four people: two spotters positioned at either end of a low-traffic and low-speed road, one responsible for placing objects on the road, and a driver.
The spotters would signal when it was safe to place the objects, and once positioned, the driver would begin recording the scene.
If one of the spotters reported an approaching vehicle, the data recording was stopped, and the object was immediately removed.
This was done to avoid disrupting the normal flow of traffic. 
The vehicle was driven at a maximum speed of 50 km / h during data recording.
Figure \ref{fig:data_monitor} shows a computer monitor placed on the road during sensor data logging.
Some of the objects placed on the road are also shown in Figure \ref{fig:objects}.

\PAR{Postprocessing.}
After data collection, post-processing is required to obtain vehicle poses and anonymize the images.
Point cloud registration was performed using KISS ICP \cite{vizzo2023kissicp}, a LiDAR odometry pipeline.
KISS ICP includes point cloud motion compensation, subsampling, adaptive thresholding to determine correspondences, and LiDAR pose estimation.
The calculated LiDAR pose was then exported in KITTI format as required for the labeling tool~\cite{behley2019semantickitti}.

Ethical considerations for this dataset include anonymizing camera images.
Identifiable data, such as faces and license plates, are processed using DashcamCleaner~\cite{dashcamcleaner} and DeepPrivacy2~\cite{hukkelas23DP2}.
DashcamCleaner uses a license plate detector to locate and blur license plates, while DeepPrivacy2 detects facial features and generates new, unidentifiable faces to replace the original ones.
Examples of anonymization can be found in the supplementary material.

\subsection{Labeling}
We define anomalies as objects that may pose a danger to the autonomous vehicle and its passengers, especially those present on the driving surface and not present in the training dataset.
In this work, we focus on road anomalies and do not consider unusual situations such as vehicle collisions or unexpected driver behavior, and we focus on the fine-grained anomaly segmentation.
In particular, we follow previous work~\cite{pinggera2016laf,blum2021fishyscapes} and focus on objects that are not present in the training set.

We labeled the data using the SemanticKITTI labeler~\cite{behley2019semantickitti}.
To generate initial labels for the scenes, we used pseudo-labeling techniques using a pre-trained SemanticKITTI model.
We instruct our annotators to group all points belonging to inliers as a single semantic class, questionable points as unlabeled, and anomaly points as the outlier class.
We instruct our labelers that it is very important to label inliers and outliers carefully while allowing for larger unlabeled regions where the annotator may be challenged.

\PAR{Annotation process.}
The domain gap between urban environments and SemanticKITTI street scenes is minimal, especially in regions close to the vehicle.
This similarity allows the acquisition of reliable initial estimates.
The annotation process was performed by three annotators, each spending eight hours a week over several months.
Each annotator spent approximately four hours on the initial labeling of each scene and an additional two hours on refining the final labels.
To ensure accuracy and consistency, the annotators cross-validated each other's work.
To further improve the quality of the labels, we examined the predictions of the baseline methods to see if any known objects were missed.
In particular, current methods tend to be highly sensitive to objects that are present but ignored during training, such as those classified in the category ``other object''.
We annotate these objects as unlabeled and ignore them in the evaluations, as they could alter the results of the anomaly segmentation methods.

\PAR{Differences to SemanticKITTI.}
In contrast to SemanticKITTI, which provides labels up to a range of $50$ meters, we labeled all points visible to the LiDAR sensor.
This was facilitated by using a LiDAR with a higher vertical resolution.
The maximum distance resulting from an anomaly in our dataset is $150$ meters, allowing for comprehensive evaluations.
However, since we train on the SemanticKITTI dataset, which contains only labels up to $50$ meters, we limit our evaluations to the $50$ meter range.
The sensor setup has $128$ LiDAR beams, compared to $64$ beams in SemanticKITTI.

\subsection{Dataset analysis}
\input{figures/5_dataset_stats/figure}

We recorded $70$ sequences with anomalies, which we fully annotated for anomaly segmentation.
Two of these labeled sequences were captured in a naturalistic manner, while the remaining $68$ were staged to include various anomalies. 
We designated $19$ sequences as a validation set and $51$ as a closed test set for benchmarking purposes.
In addition, we recorded and annotated two sequences with no anomalies for training and validation of the inlier performance ``STU-inlier'', to reduce the domain gap between our dataset and SemanticKITTI.

Our dataset defines two primary classes: \textit{inliers} and \textit{outliers}. 
\textit{Inliers} encompasses classes that are used to train a model, and contains objects expected on the road, such as vehicles, pedestrians, and road infrastructure.
\textit{Outliers}, on the other hand, refer to out-of-distribution elements or anomalies that are atypical or unexpected in the driving scene, shown in Figure~\ref{fig:objects}.
One additional class, \textit{unlabeled} is ignored in the evaluation and typically includes classes that the model has seen during training, but was not supervised in, such as parking meters, utility boxes, and lamps.
We provide anomaly annotations at both the instance and semantic levels, enabling a comprehensive evaluation of models for semantic and instance segmentation in 3D environments.

Each outlier object in our dataset has an instance label unique for a sequence.
Figure~\ref{fig:instance_properties} presents histograms that display the distribution of key properties for these instances, including the average and maximum number of LiDAR points per instance, instance heights, and the number of instances per scan.
This analysis provides insights into the structural characteristics and variability of data within point clouds.
Specifically, Figure~\ref{fig:instance_properties} indicates that most instances have very few LiDAR points on average, but a sequence may have multiple instances per sequence.

Furthermore, Figure~\ref{fig:xy_distr} presents a scatter plot of detected anomalies within the vehicle frame, highlighting a higher density of anomalies along the longitudinal ($X$) axis both in front and behind the vehicle.
The accompanying histograms depict the distribution of the anomalies along each axis, noting that the anomalies are identified at distances ranging up to 150 meters in front of the vehicle.
Figure~\ref{fig:anomaly-distance} explores the relationship between the distance to an anomaly and its size, showing that the proportion of LiDAR point cloud obstacles decreases as the distance increases.
This trend is due to the radial nature of LiDAR, which has lower vertical resolution at greater ranges.
For benchmarking purposes, we focus exclusively on instances with more than five LiDAR points, highlighted in blue, ensuring that the evaluation targets detectable anomalies.

\subsection{Evaluated Baselines}
The formulation of the anomaly segmentation task restricts training to avoid direct supervision of OOD objects.
Generally, OOD objects are not present in the training set, but could be present during inference~\cite{gasperini2023u3hs}.
Since there are no methods that could be applied directly for our setup, we adopt several baseline methods from the 2D anomaly segmentation to 3D LiDAR segmentation.
First, we evaluate simple baselines, such as Max-Logit~\cite{hendrycks2018baseline}, Monte-Carlo Dropout~\cite{srivastava2014mcdropout}, and Deep Ensembles~\cite{lakshminarayanan2017deepensemble}.
Secondly, we train one of the models to predict ignored regions during training, which could be used to classify ``void'' points during testing~\cite{blum2021fishyscapes}.
We also adapt RbA~\cite{nayal2023rba}, one of the state-of-the-art methods for image anomaly segmentation, to the 3D LiDAR segmentation method Mask4Former-3D~\cite{yilmaz2024mask4former}.

\input{tables/0_main}

\PAR{Deep Ensembles, MC Dropout, and MaxLogit.}
Deep Ensembles~\cite{lakshminarayanan2017deepensemble} and MC Dropout~\cite{srivastava2014mcdropout} are very common methods that practitioners use due to ease of implementation and application.
For Deep Ensembles, we train three Mask4Former-3D~\cite{yilmaz2024mask4former} models.
For MC dropout, we train one Mask4Former-3D model with enabled dropout in the transformer decoder for self- and cross-attention operations, and during inference we repeat the transformer decoder operation 50 times.
As a simple baseline that uses only one model and a single forward pass, we use the MaxLogit~\cite{hendrycks2018baseline} method.

\PAR{RbA}~\cite{nayal2023rba} is a method that uses the ability of MaskFormer-Style models~\cite{cheng2021maskformer,cheng2021mask2former} to predict multiple overlapping segmentation masks.
MaskFormer models have an option to disable queries and might not predict any masks if no objects are present.
Points for which no masks were predicted will have the highest anomaly score.
The Mask4Former model follows the same architectural paradigm and has a query deactivation mechanism that allows for no prediction in uncertain regions.
This allows us to use the RbA method to generate anomaly scores within our setup.
Unlike the original RbA model setup, we did not modify the transformer decoder. 

\PAR{Void Classifier}
We follow a common baseline~\cite{blum2021fishyscapes} by treating the unlabeled regions as an additional semantic class during training.

%% file: figures/3_data_collection/figure.tex
\begin{figure}[t]
\centering
\begin{subfigure}[]{\columnwidth}
    \centering
    \begin{tikzpicture}[spy using outlines={red, magnification=2.5, height=1.5cm, width=2cm, connect spies}]
        \node {\includegraphics[trim={0cm 2.5cm 0cm 10cm},clip,width=\columnwidth]{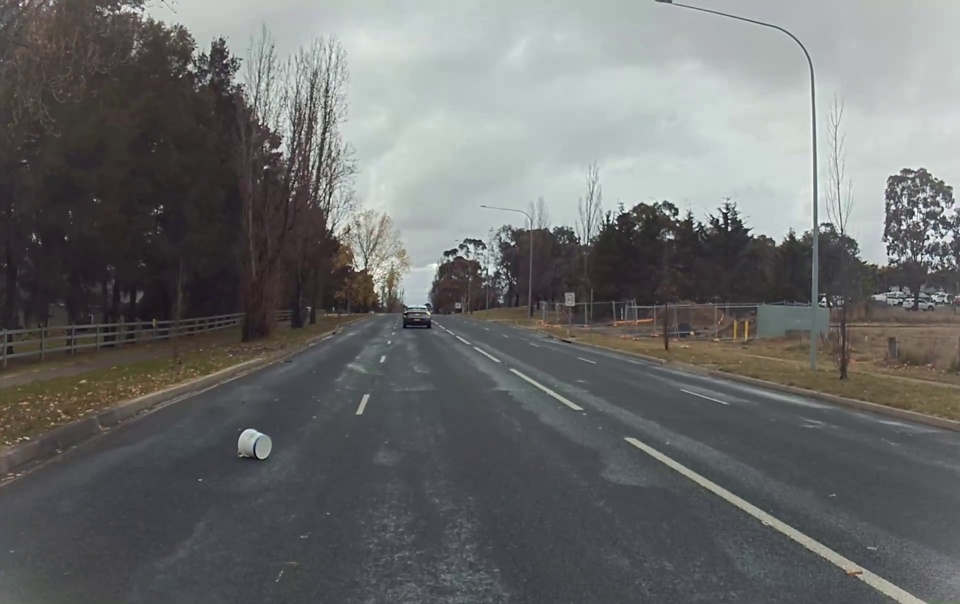}};
        \spy on (-2,-0.3) in node [right] at (0.6,0);
    \end{tikzpicture}
    \caption{\small Naturalistic data collection.}
    \label{fig:data_naturalistic}
\end{subfigure}

\centering
    
\begin{subfigure}[b]{\columnwidth}
    \centering
    \begin{tikzpicture}[spy using outlines={red, magnification=2.5, height=1.5cm, width=2cm, connect spies}]
        \node {\includegraphics[trim={0cm 2.5cm 0cm 10cm},clip,width=\columnwidth]{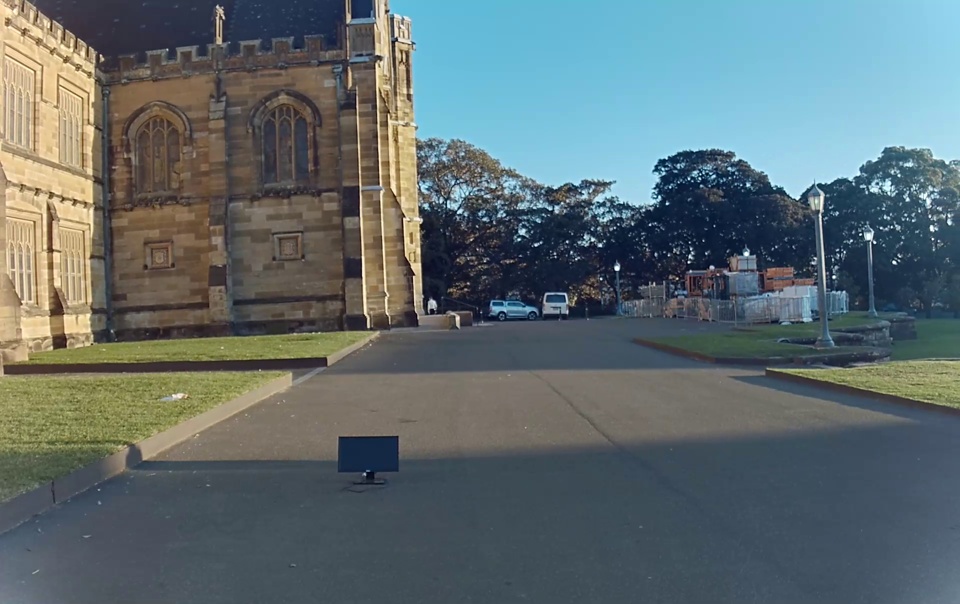}};
        \spy on (-1,-0.45) in node [right] at (0.6,0);
    \end{tikzpicture}
    \caption{\small Data collection in controlled environment.}
    \label{fig:data_monitor}
\end{subfigure}

\caption{\small 
Data collection conducted in a naturalistic manner (a) and controlled environment (b) with objects on the road.
}
\label{fig:data_collection}
\end{figure}

%% file: figures/2_objects/figure.tex
\begin{figure*}[ht]
\centering
\begin{subfigure}[]{0.16\textwidth}
\centering
\includegraphics[trim={1.2cm 1cm 2cm 1cm},clip,width=\textwidth]{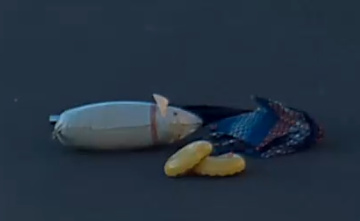}
\end{subfigure}
\begin{subfigure}[]{0.16\textwidth}
\centering
\includegraphics[trim={2.2cm 1cm 1cm 1cm},clip,width=\textwidth]{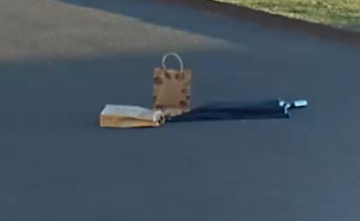}
\end{subfigure}
\begin{subfigure}[]{0.16\textwidth}
\centering
\includegraphics[trim={2.2cm 1.5cm 1cm 0.5cm},clip,width=\textwidth]{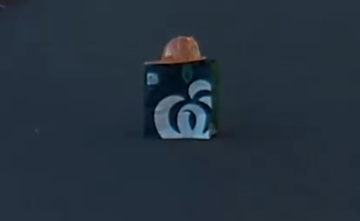}
\end{subfigure}
\begin{subfigure}[]{0.16\textwidth}
\centering
\includegraphics[width=\columnwidth]{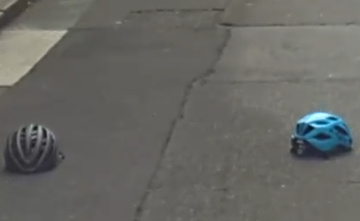}
\end{subfigure}
\begin{subfigure}[]{0.16\textwidth}
\centering
\includegraphics[trim={1.7cm 1cm 1.7cm 1cm},clip,width=\columnwidth]{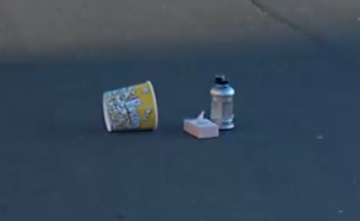}
\end{subfigure}
\begin{subfigure}[]{0.16\textwidth}
\centering
\includegraphics[width=\columnwidth]{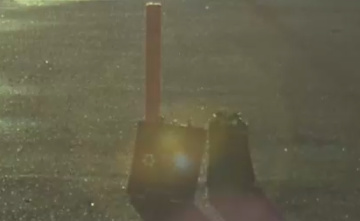}
\end{subfigure}

\begin{subfigure}[]{0.16\textwidth}
\centering
\includegraphics[width=\textwidth]{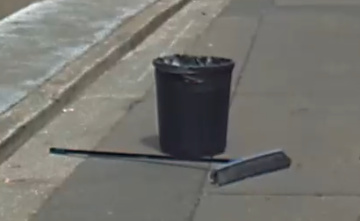}
\end{subfigure}
\begin{subfigure}[]{0.16\textwidth}
\centering
\includegraphics[width=\textwidth]{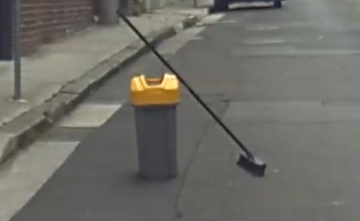}
\end{subfigure}
\begin{subfigure}[]{0.16\textwidth}
\centering
\includegraphics[width=\textwidth]{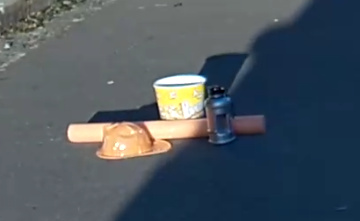}
\end{subfigure}
\begin{subfigure}[]{0.16\textwidth}
\centering
\includegraphics[trim={2.2cm 1cm 2.5cm 2cm},clip,width=\columnwidth]{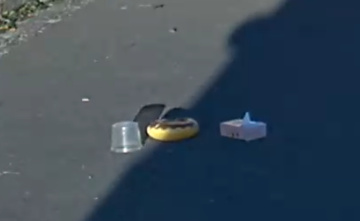}
\end{subfigure}
\begin{subfigure}[]{0.16\textwidth}
\centering
\includegraphics[width=\columnwidth]{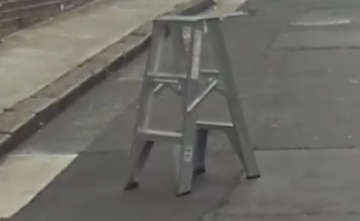}
\end{subfigure}
\begin{subfigure}[]{0.16\textwidth}
\centering
\includegraphics[width=\columnwidth]{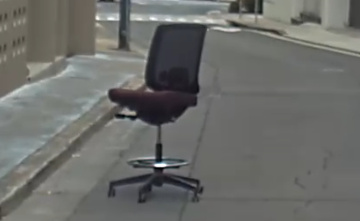}
\end{subfigure}

\begin{subfigure}[]{0.16\textwidth}
\centering
\includegraphics[width=\textwidth]{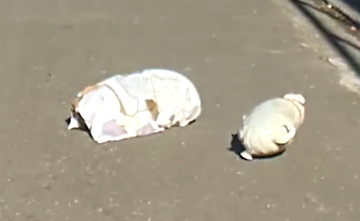}
\end{subfigure}
\begin{subfigure}[]{0.16\textwidth}
\centering
\includegraphics[width=\textwidth]{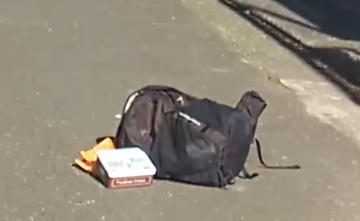}
\end{subfigure}
\begin{subfigure}[]{0.16\textwidth}
\centering
\includegraphics[trim={2.2cm 1.3cm 2.5cm 1.7cm},clip,width=\textwidth]{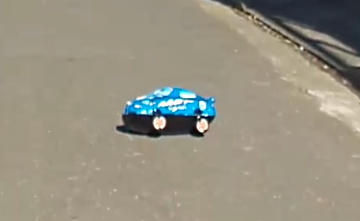}
\end{subfigure}
\begin{subfigure}[]{0.16\textwidth}
\centering
\includegraphics[trim={0 0.15cm 0 0},clip,width=\columnwidth]{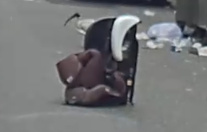}
\end{subfigure}
\begin{subfigure}[]{0.16\textwidth}
\centering
\includegraphics[trim={0 0.15cm 0 0},clip,width=\columnwidth]{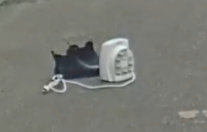}
\end{subfigure}
\begin{subfigure}[]{0.16\textwidth}
\centering
\includegraphics[trim={0 0.25cm 0 0},clip,width=\columnwidth]{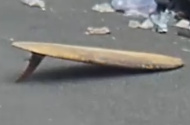}
\end{subfigure}

\begin{subfigure}[]{0.16\textwidth}
\centering
\includegraphics[trim={1.0cm 0.0cm 1.3cm 0cm},clip,width=\textwidth]{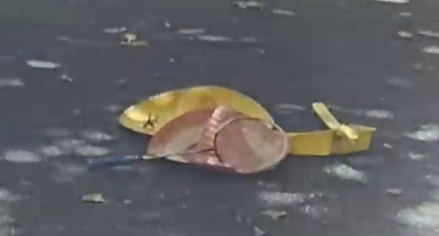}
\end{subfigure}
\begin{subfigure}[]{0.16\textwidth}
\centering
\includegraphics[trim={0.0cm 0.0cm 0.25cm 0cm},clip,width=\textwidth]{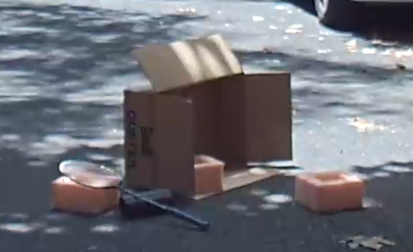}
\end{subfigure}
\begin{subfigure}[]{0.16\textwidth}
\centering
\includegraphics[trim={1.0cm 0.0cm 1.0cm 0cm},clip,width=\textwidth]{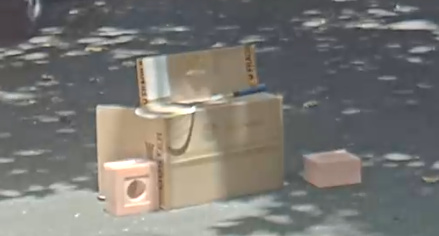}
\end{subfigure}
\begin{subfigure}[]{0.16\textwidth}
\centering
\includegraphics[trim={1.0cm 0.0cm 1.0cm 0cm},clip,width=\columnwidth]{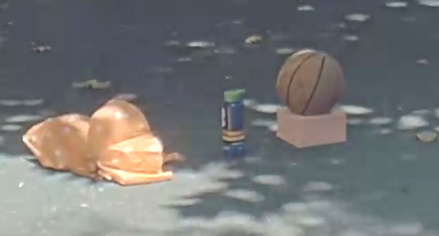}
\end{subfigure}
\begin{subfigure}[]{0.16\textwidth}
\centering
\includegraphics[trim={1.0cm 0.0cm 1.0cm 0cm},clip,width=\columnwidth]{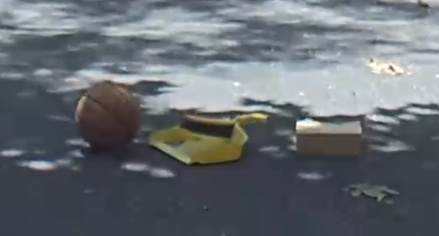}
\end{subfigure}
\begin{subfigure}[]{0.16\textwidth}
\centering
\includegraphics[trim={1.0cm 0.0cm 1.0cm 0cm},clip,width=\columnwidth]{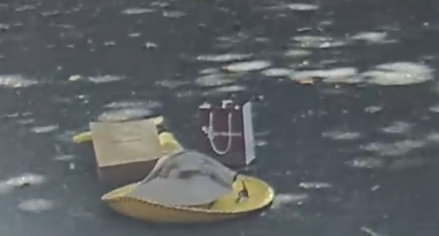}
\end{subfigure}

\caption{\small 
\textbf{Different anomalies in the STU dataset.}
Different objects on the road used for staged data collection.
We pick objects such that have no intersection with the inlier dataset and place them on roads in different locations and illumination conditions.
Objects might touch each other, be very small, as large as a chair or a surf board, and could cause an accident if a car would drive over them.
}
\label{fig:objects}
\end{figure*}

%% file: figures/5_dataset_stats/figure.tex
\begin{figure*}%
    \centering
    \begin{subfigure}{0.5\columnwidth}
    \includegraphics[trim={0 0 0 0},clip,width=\columnwidth]{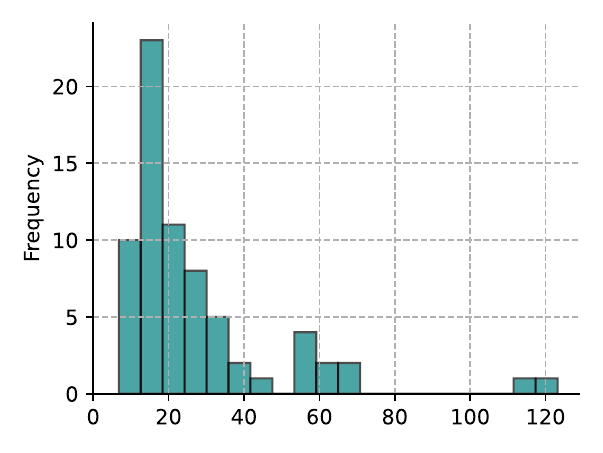}
    \caption{\small Avg. num of points/sequence.}
    \label{fig:dist_avg}
\end{subfigure}
\begin{subfigure}{0.5\columnwidth}
    \includegraphics[trim={0 0 0 0},clip, width=\columnwidth]{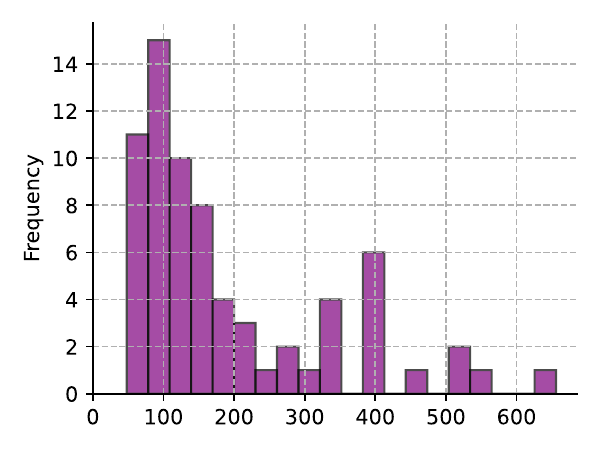}
    \caption{\small Max. points per instance.}
    \label{fig:dist_points}
\end{subfigure}
\begin{subfigure}{0.5\columnwidth}
    \includegraphics[trim={0 0 0 0},clip,width=\columnwidth]{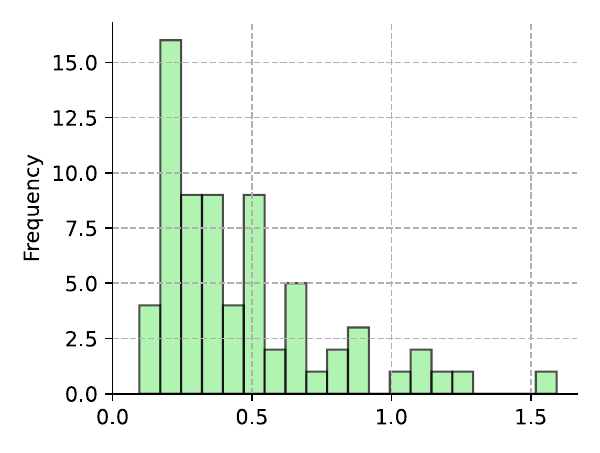}
    \caption{\small Max. height of instances (m).}
    \label{fig:dist_height}
\end{subfigure}
\begin{subfigure}{0.5\columnwidth}
    \includegraphics[trim={0 0 0 0},clip,width=\columnwidth]{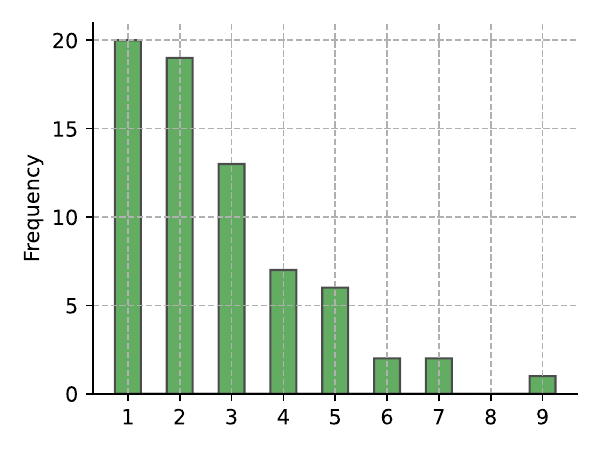}
    \caption{\small Numb. of instances/sequence.}
    \label{fig:dist_inst_scan}
\end{subfigure}

    \vspace{-0.75em}
    \caption{
    \small \textbf{Anomaly Instance Properties}.
    A typical recorded anomaly has on average less then $50$ points per sequence (a), with less then $300$ (b) points at maximum, and a maximum height below one meter (c).
    We record up to nine individual anomaly instances in the same sequence (d).
    }
    \label{fig:instance_properties}
\end{figure*}

\begin{figure}
\begin{tikzpicture}
\node[anchor=south west, inner sep=0] (image) at (0,0) {\includegraphics[width=\columnwidth]{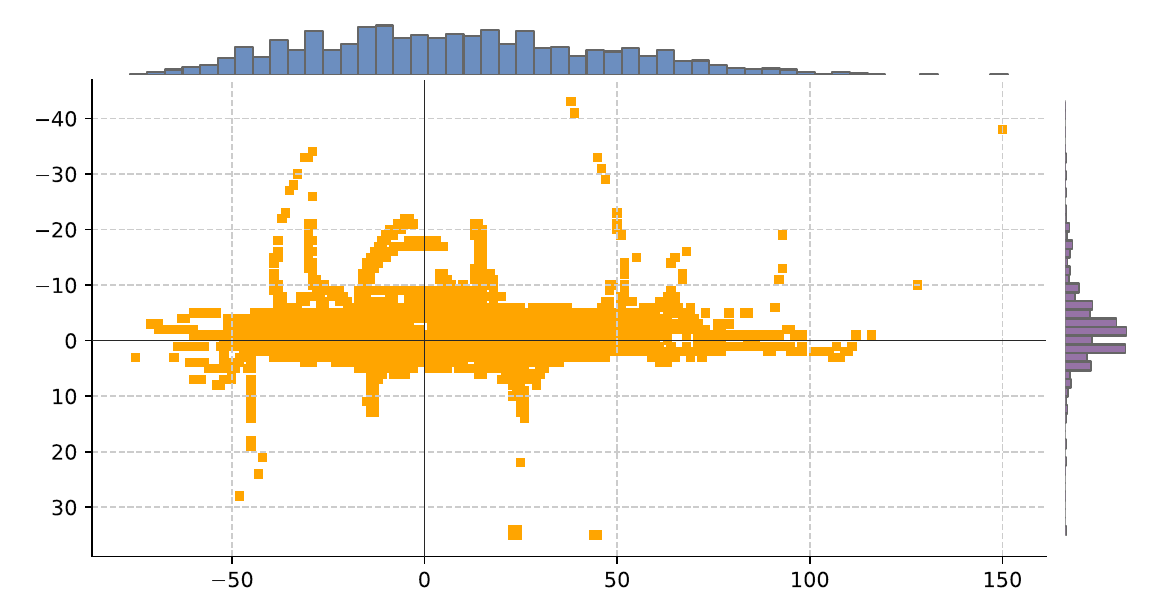}};
\node[below, yshift=0mm, xshift=-3mm] at (image.south) {\footnotesize X (Longitudinal)};
\node[rotate=90, left, yshift=-1mm, xshift=6mm, anchor=base east] at (image.west) {\footnotesize Y (Lateral)};
\end{tikzpicture}
\vspace{-2em}
\caption{
\small 
Distribution of anomalies along the vehicle's reference frame.
Most of the points appear around the vehicle.
}
\label{fig:xy_distr}
\end{figure}

\begin{figure}
    \centering
    \begin{tikzpicture}
        \begin{axis}[
            xlabel={\small Number of Points for an Anomaly Instance},
            ylabel={\small Distance to Anomaly Instance (meters)},
            width=0.8\linewidth,
            height=0.6\linewidth,
            scale only axis,
            grid=both,
            minor tick num=3,
            xmax=1000,
            ymax=150,
            ymajorgrids=true,
            xmajorgrids=true,
            log ticks with fixed point,
            xmode=log,
            ymode=log,
            ytick={1, 10, 100, 1000},
            yticklabels={1, $10^1$, $10^2$, $10^3$},
            legend style={at={(0.6,0.95)}, anchor=north west},
        ]
            \addplot[
                only marks, 
                opacity=0.03, 
                mark size=1.5pt, 
                color=m_red_border, %
            ] table[col sep=comma] {figures/5_dataset_stats/rest_of_data.csv};

            \addplot[
                only marks, 
                opacity=0.5, 
                mark size=1.5pt, 
                color=m_blue_border %
            ] table[col sep=comma] {figures/5_dataset_stats/lower_right_corner.csv};

            \addplot[red, very thick, dashed] coordinates {(5, 50) (1000, 50)};
            \addplot[red, very thick, dashed] coordinates {(5, 1) (5, 150)};

        \end{axis}
    \end{tikzpicture}
    \caption{\small
    We follow SemanticKitti's maximum distance threshold and evaluate instances that are within a $50$ meter radius of the vehicle.
    In addition, we constrain evaluation of point and object level metrics to a minimum of 5 points.
    }
    \label{fig:anomaly-distance}
\end{figure}
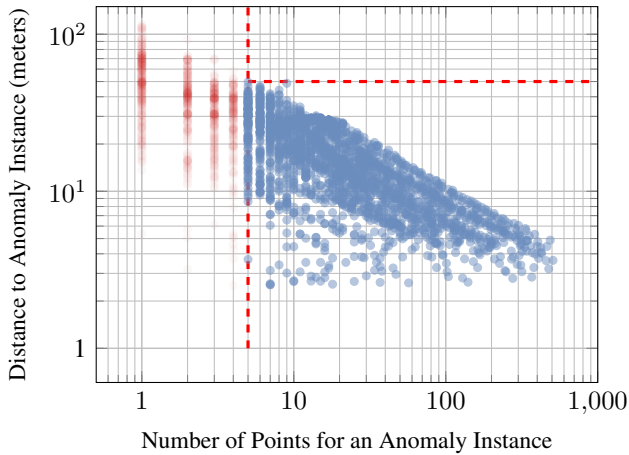

%% file: tables/0_main.tex
\begin{table*}[t]
    \centering
    \caption{\small Anomaly Segmentation Performance of the selected baseline methods.}
    \resizebox{\textwidth}{!}{
    \begin{tabular}{lccccccccccc}
        \toprule
        \multirow{2}{*}{Method} & \multirow{2}{*}{OoD Data} & \multicolumn{3}{c}{Point-Level OOD} && \multicolumn{5}{c}{Object-Level OOD} \\
        \cline{3-5}
        \cline{7-11}
         &  & AUROC~$\uparrow$ & FPR@95~$\downarrow$ & AP~$\uparrow$ && RecallQ & SQ & RQ & UQ & PQ \\
        \midrule
        Deep Ensemble~\cite{lakshminarayanan2017deepensemble}         & \xmark  & \textbf{86.74} & \textbf{58.05} & \cellcolor{gray!20}{\textbf{5.17}} && 16.75 & \ul{84.49} & \textbf{10.43} & 14.16 & \cellcolor{gray!20}{\textbf{8.81}} \\
        MC Dropout~\cite{srivastava2014mcdropout}       & \xmark  & 61.51 & 82.37 &  \cellcolor{gray!20}{0.11} && 
        2.25 & \textbf{86.72} & 1.95 & 2.14 & \cellcolor{gray!20}{1.86} \\
        Max Logit~\cite{hendrycks2018baseline}        & \xmark  & 84.53 & 81.49 & \cellcolor{gray!20}{0.95} && \textbf{26.14} & 83.06 & 2.13 & \textbf{21.71} & \cellcolor{gray!20}{1.77} \\
        Void Classifier~\cite{blum2021fishyscapes}  & \cmark  & \ul{85.99} & \ul{78.60} & \cellcolor{gray!20}{\ul{3.92}} && 17.64 & 84.40 & \ul{8.19} & 14.89 & \cellcolor{gray!20}{\ul{6.91}} \\
        RbA~\cite{nayal2023rba}             & \xmark  & 66.38 & 100.0 & \cellcolor{gray!20}{0.81} && \ul{24.04} & 83.28 & 3.23 & \ul{20.02} & \cellcolor{gray!20}{2.69} \\
        \bottomrule
    \end{tabular}}
    \label{tab:anomaly_segmentation_performance_difficulty}
\end{table*}

\begin{table}[t]
    \centering
    \caption{
    \small Closed-set performance of the methods on validation sets of SemanticKITTI~\cite{behley2019semantickitti} and STU-inlier dataset.
    $^*$ denotes a model trained only on SemanticKITTI dataset.
    }
    \resizebox{\linewidth}{!}{
    \begin{tabular}{lcccccccccc}
        \toprule
        \multirow{2}{*}{Method} & \multicolumn{3}{c}{SemanticKITTI} && \multicolumn{3}{c}{STU-inlier Dataset} \\
        \cline{2-4}
        \cline{6-8}
         & PQ & SQ & RQ && PQ & SQ & RQ \\
        \midrule 
        Mask-PLS$^*$~\cite{marcuzzi2023maskpls} & 59.9 & 76.4 & 69.1 && 39.6 & 67.0 & 50.5 \\
        Mask-PLS~\cite{marcuzzi2023maskpls} & 55.6 & 76.0 & 64.5 && \ul{50.9} & 69.4 & 62.2 \\
        Mask4Former-3D$^*$~\cite{yilmaz2024mask4former} & \textbf{61.9}  & 81.3 & 71.6 && 42.8 & 70.6 & 53.2 \\
        Mask4Former-3D~\cite{yilmaz2024mask4former} & \ul{60.7} & 81.0 & 70.4  && \textbf{52.7} & 73.8 & 64.7   \\
        \bottomrule 
    \end{tabular}}
    \label{tab:inlier_metrics}
\end{table}

%% file: sec/4_experiments.tex
\section{Baseline Evaluation}
\PAR{Training data.}
We use SemanticKITTI~\cite{behley2019semantickitti}, Panoptic-CUDAL~\cite{tseng2025panopticcudal}, and an additional recorded sequence from to jointly train our models.
The Panoptic-CUDAL dataset, contains six sequences for training and one for validation, and was mostly recorded in rural locations, while our dataset consists mainly of urban scenes.
To reduce the domain gap, we have recorded and annotated two additional inlier scenes from an urban setting ``STU-inlier'', one for training and one for evaluation.
These training sequences closely resemble the environments in which anomaly scenes were recorded.
We train models with additional data recorded from the same vehicle setup as Panoptic-CUDAL and labeled the data with $19$ SemanticKITTI labels.
However, SemanticKITTI and our recorded dataset contain "other-object" that acts as a miscellaneous class for all objects that do not fit other-class definitions.
A typical object is a bench, a garbage bin, or an advertisement post.
To strictly define anomalies, we analyze all other-objects present in our training set, as well as the "Movable\_Object.Debris" and "Pushable.Pullable" class objects in the NuScenes training set.
We specifically design our dataset such that we do not have an intersection between anomalous objects and the aforementioned classes.
Our Void Classifier training procedure consists of training on this unlabeled class that is present in the training data.

\PAR{Models.}
We train MaskPLS~\cite{marcuzzi2023maskpls} and Mask4Former~\cite{yilmaz2024mask4former} models jointly on both SemanticKitti and our training sequences. %
These models achieve state-of-the-art results on the SemanticKITTI~\cite{behley2019semantickitti} dataset for single-scan and multi-scan settings.
However, while our dataset supports evaluation of temporal sequences, we performed our evaluation and training on single scans only.
In our experiments, we refer to Mask4Former trained on single scans as Mask4Former-3D.
All anomaly segmentation methods use Mask4Former-3D as the main model.

\PAR{Metrics.}
To evaluate the performance of our methods, we use common metrics in the anomaly segmentation community in 2D and panoptic segmentation in 3D LiDAR. %
Following SemanticKitti, we evaluated both inlier and outlier performance on points within $50$ meters of the vehicle, and on objects with at least five points.

\PAR{Point-Level Metrics.}
To evaluate the performance of the anomaly segmentation at the point level, we use the common metrics of Average Precision (AP), False-Positive Rate at $95\%$ True-Positive Rate (FPR95), and Area Under Receiver Operating characteristic Curve (AUROC).
Commonly used in the anomaly segmentation community~\cite{blum2021fishyscapes}.

\PAR{Object-Level Metrics.}
To be consistent with the panoptic segmentation setup, we use panoptic metrics to evaluate the model performance on inlier classes.
For the inlier data, we compute the Panoptic Quality (PQ) on the SemanticKitti classes.
We report detailed class scores in the Supplementary.
In contrast to anomaly segmentation, the open task formulation emphasizes the recall of all objects. %
Formally, the Unknown Quality (UQ) metric is used in such a setting~\cite{wong2019osis}.
However, in the case of anomaly segmentation, since anomalies can pose serious threats to the vehicle, it is important to predict as few false positives as possible, while also producing accurate segmentation labels for anomaly objects.
Panoptic Quality uses an F-score and penalizes the prediction of multiple false positives.
Thus, for anomaly segmentation baselines, we report both Panoptic Quality and Unknown Quality on a single class.
For inlier performance, we report the Panoptic Quality score on the SemanticKITTI classes.

Panoptic Quality~\cite{kirillov2019panoptic} is the geometric mean between Recognition Quality (RQ) and Segmentation Quality (SQ) for a class $c$:
\newline

\resizebox{0.9\columnwidth}{!}{
$
\text{PQ}_c = \underbrace{\frac{\sum_{(p, g) \in TP_c} \text{IoU}(p, g)}{|TP_c|}}_{\text{Segmentation Quality (SQ)}} 
\times 
\underbrace{\frac{|TP_c|}{|TP_c| + \frac{1}{2} |FP_c| + \frac{1}{2} |FN_c|}}_{\text{Recognition Quality (RQ)}}
$}

\vspace{5mm}
where $TP$, $FP$, and $FN$ are True Positives, False Positives, and False Negatives for objects with Intersection over Union (IoU) higher than $50\%$.
The final PQ metric is calculated as an average over all classes $c$.
Ignore points are removed from the scene prior to evaluation and erroneous predictions in the ignore region are not penalized.

To evaluate the recall of anomaly objects, we use the Unknown Quality (UQ) metric~\cite{wong2019osis}:
\begin{equation}
\text{UQ} = \underbrace{\frac{\sum_{(p, g) \in TP} \text{IoU}(p, g)}{|TP|}}_{\text{Segmentation Quality (SQ)}} 
\times 
\underbrace{\frac{|TP|}{|TP| + |FN|}}_{\text{Recall Quality (RecallQ)}}.
\nonumber
\end{equation}
The metric does not penalize $\mathit{FP}$ and allows for the evaluation of object recall.
However, the metric counts objects as $\mathit{TP}$ only if they have an IoU of at least $50\%$ with ground truth.

\subsection{Experiments}
\PAR{Inlier Performance on SemanticKitti and STU-inlier dataset.}
We train models jointly with SemanticKITTI~\cite{behley2019semantickitti}, Panoptic-CUDAL~\cite{tseng2025panopticcudal} and our additional sequence, and evaluate the SemanticKITTI and STU-inlier validation sets.
The jointly trained model demonstrates improved performance on our STU-inlier dataset compared to models trained solely on the SemanticKITTI dataset, suggesting a small domain gap between our dataset and SemanticKITTI.
Evaluation on the proposed STU-inlier sequence shows good segmentation performance for inlier classes that is critical for anomaly segmentation.
The metrics for the performance of the baseline method in the validation sets are provided in Tab.~\ref{tab:inlier_metrics}.

\PAR{Out-of-Distribution (OOD) Performance.}
We evaluate baseline methods for anomaly segmentation at both the point and object levels.
Our findings indicate that directly transferring methods from 2D to 3D does not yield similar performance.
Our point clouds typically contain around $100,000$ points, with only a few dozen being anomalous, as shown in Fig.~\ref{fig:anomaly-distance}. 
Due to the significant class imbalance, point-level scores are generally low, highlighting the difficulty of the task.
A high FPR@95, as noted in ODIN~\cite{hsu2020odin}, suggests that models often misclassify road objects as inlier classes.
Since LiDAR point clouds have lower resolution at a distance, predicting a small anomaly object consisting of a few points is a hard task.
It is common for large anomalous objects to be predicted as pedestrians or other vehicles, as depicted in Fig.~\ref{fig:failure_case}.
Anomalous objects frequently occur in familiar contexts, which leads the model to confidently assign inlier labels to such anomalies in many cases.
This behavior contrasts significantly with models trained on 2D data, where a simple max logit approach~\cite{Hendrycks2022ScalingOD} outperforms Deep Ensembles~\cite{chan2021segmentmeifyoucan}.
However, the Deep Ensemble model stands out by producing fewer false positives at the point level, resulting in a lower FPR@95 and a higher PQ metric.
Despite this reduction in false positives, the average precision (AP) remains low compared to 2D datasets.
We attribute this to the high confidence with which individual models predict inlier objects, with disagreements between models contributing to the improved FPR@95.
We refer readers to the supplementary material for a more detailed investigation of the results.

%% file: sec/5_conclusion.tex
\input{figures/4_failure_case/figure}
\section{Conclusions and Future Work}
In this paper, we introduced the STU dataset, designed for anomaly segmentation in LiDAR point clouds.
The dataset was collected using eight cameras and a LiDAR sensor, which provides a surround view of environments through images and high-resolution LiDAR data.
We extensively annotated point clouds with ``unlabeled'', ``anomaly'', and ``inlier'' classes, as well as introduced additional sequences without anomalies for training and evaluation.
The results of the baseline methods indicate that there is a significant performance gap between 2D and 3D anomaly segmentation, which underscores numerous ongoing research opportunities in the field. 
Our objective with this paper is to provide researchers with annotated data and standardized benchmarks to facilitate the comparison of their anomaly segmentation approaches in 3D.
We believe the dataset could be especially useful for the evaluation of methods that use temporal LiDAR data and multi-modal data in the future, and can drive innovation in developing more robust anomaly detection systems that can better handle complex real-world scenarios.

\PAR{Acknowledgments.}{ \small
A. Nekrasov’s acknowledges funding by BMBF project ``WestAI'' (grant no. 01IS22094D).
J. S. Berrio Perez acknowledges funding by the German Academic Exchange Service (DAAD Project-ID 30001831).
}

%% file: figures/4_failure_case/figure.tex
\begin{figure}[t]
    \centering
    \includegraphics[trim={0 0 0 0},clip, width=0.45\columnwidth]{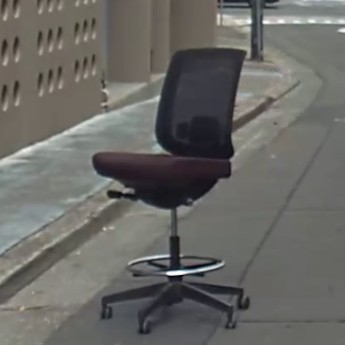}
    \includegraphics[trim={0 0 0 0},clip, width=0.45\columnwidth]{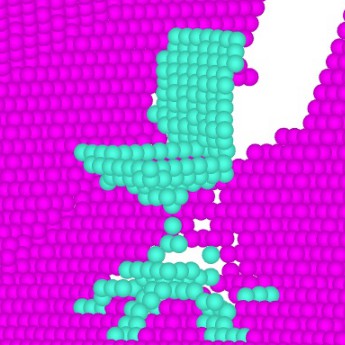}
    \includegraphics[trim={0 0 0 0},clip, width=0.45\columnwidth]{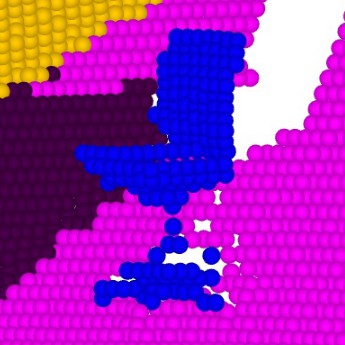}
    \includegraphics[trim={0 0 0 0},clip, width=0.45\columnwidth]{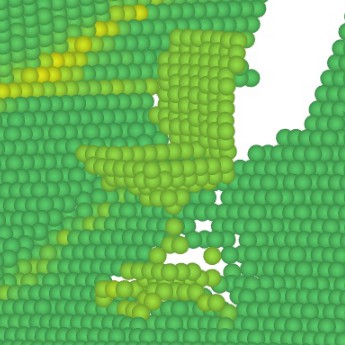}
    \caption{
    \small Example of a failure case anomaly segmentation.
    For the chair on the road, labeled as anomaly in upper left, model predicts ``other-vehicle'' \protect\colorsquare{blue} class, in lower left, with a high certainty, as indicated by MaxLogit scores on the bottom right.
    }
    \label{fig:failure_case}
\end{figure}

%% file: sec/X_suppl.tex
\clearpage
\setcounter{page}{1}
\maketitlesupplementary

\section{Hardware Setup}
The sensors and hardware included in the data collection platform are as follows:
\begin{itemize}
    \item 5 SF3325 automotive GMSL cameras (ONSEMI CMOS image sensor AR0231), SEKONIX ultra high-resolution lens with 60 horizontal and 38 vertical FOV, images captured at a resolution of 1928 $\times$ 1208 (2.3M pixel) at 30 Hz.
    \item 3 SF3324 automotive GMSL cameras, (ONSEMI CMOS image sensor AR0231), SEKONIX ultra high-resolution lens with 120 horizontal and 73 vertical FOV, images captured at a resolution of 1928 $\times$ 1208 (2.3M pixel) at 30 Hz.
    \item 1 OS1-128 Ouster Lidar, with a vertical resolution of 128 beams within a 45 FOV and range of 200 meters, point cloud captured at 10 Hz.
    \item 1 NVIDIA DRIVE Pegasus, with two NVIDIA Xavier™ SoCs.
\end{itemize}

\begin{figure}[h!]
    \centering
    \includegraphics[trim={2.2cm 0cm 2.8cm 0cm},clip,width=\linewidth]{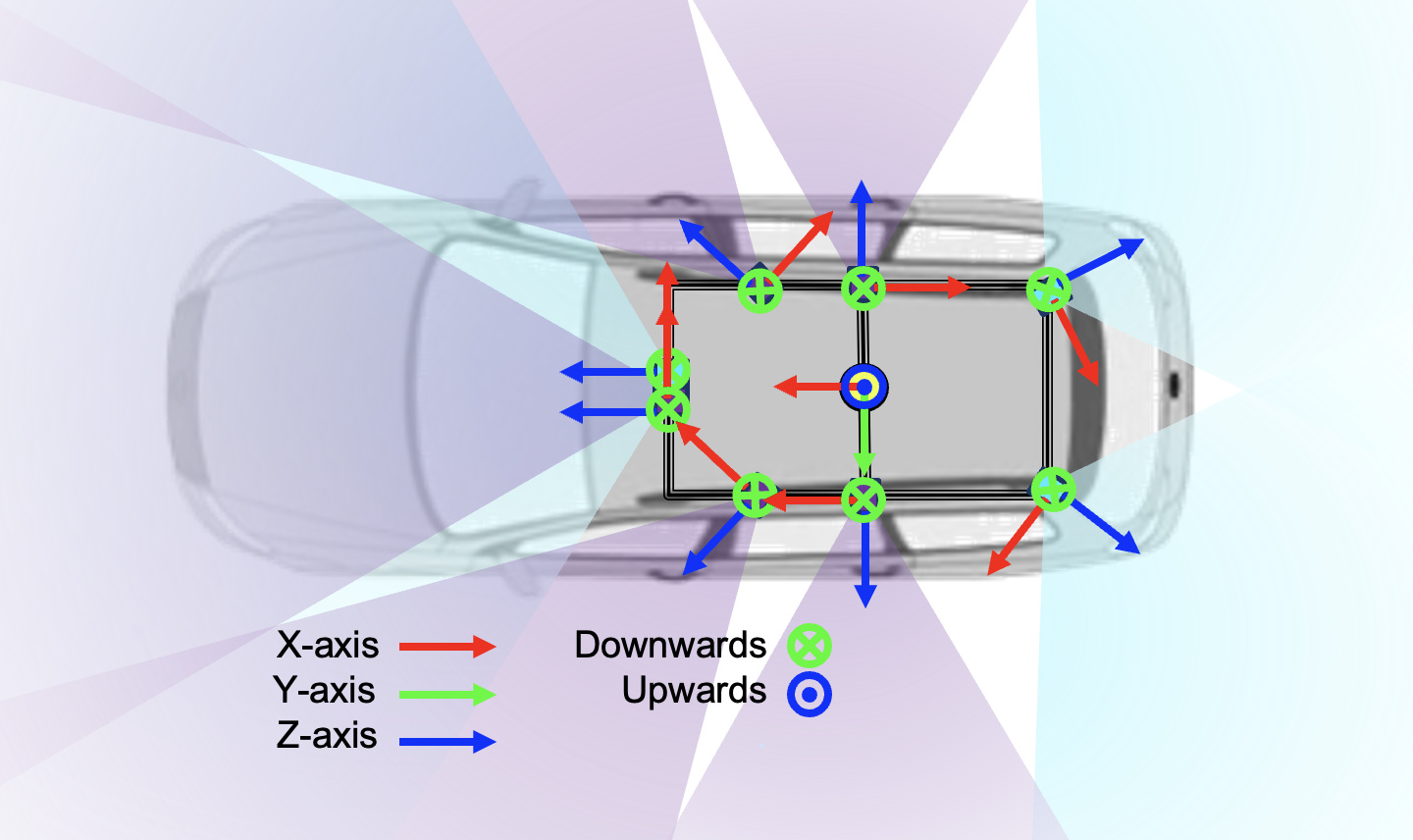}
    \caption{\small Sensor setup of the data collection vehicle.
    The field of view for the 60-degree and 120-degree cameras is represented in purple and blue, respectively. }
    \label{fig:car_frames}
\end{figure}

The placement and reference frames of the sensors on the vehicle are shown in Figure \ref{fig:car_frames}.
The arrangement of the cameras and LiDAR sensors allows the vehicle to achieve a full 360$^\circ$ field of view (FOV) of its surroundings.

\subsection{Extrinsic Calibration}
\begin{figure}[h!]
    \centering
    \includegraphics[trim={0cm 0cm 0cm 0cm},clip,width=\linewidth]{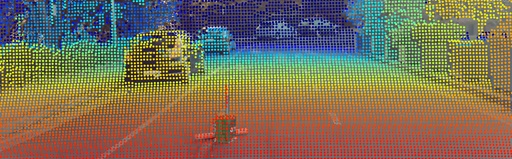}
    \caption{\small Lidar-Camera projection. The point cloud is colored by distance to the camera.}
    \label{fig:projection_1}
\end{figure}

The camera positions on the vehicle were determined through a LiDAR-camera calibration process in which we computed the homogeneous transformation matrices from the LiDAR to each camera.
Without ground truth for these transformations, their accuracy is typically validated visually by examining the correspondence between objects in the camera images and the LiDAR point cloud.
In this case, Figure~\ref{fig:projection_1} shows the alignment between the LiDAR point cloud and an image captured by the front camera, illustrating the accuracy of the calibration process.

\subsection{Software}
A Robotic Operation System (ROS) framework manages the LiDAR point-cloud acquisition pipeline.
Camera data is captured using the NVIDIA DRIVE Pegasus video capture and image compression pipeline.
All images are encoded and stored as H.264 video, with associated metadata stored in a custom ROS message.

\section{Data Collection}

\begin{figure}[b!]
    \centering
    \begin{subfigure}[]{0.32\columnwidth}
    \centering
    \includegraphics[trim={0cm 0cm 0cm 0cm},clip,width=\columnwidth]{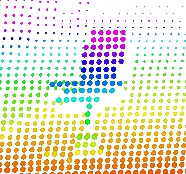}
    \caption{Chair.}
    \label{fig:data_chair}
    \end{subfigure}
    \begin{subfigure}[]{0.32\columnwidth}
    \centering
    \includegraphics[trim={0cm 0cm 0cm 0cm},clip,width=\columnwidth]{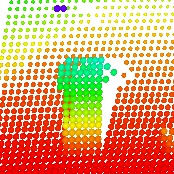}
    \caption{Garbage bin.}
    \label{fig:data_bin}
    \end{subfigure}
    \begin{subfigure}[]{0.32\columnwidth}
    \centering
    \includegraphics[trim={0cm 0cm 0cm 0cm},clip,width=\columnwidth]{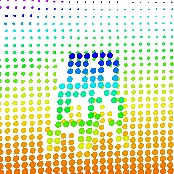}
    \caption{Ladder.}
    \label{fig:data_ladder}
    \end{subfigure}
    \caption{\small Point cloud of some objects on the road colored by height.}
    \label{fig:data_collection_1}
\end{figure}

For staged data collection, we used a diverse collection of objects, including buckets, indoor garbage bins, brooms, chairs, pots, stuffed animals, balloons, balls, backpacks, bags, pillows, shoes, umbrellas, hats, yoga mats, helmets, swimming noodles, tissue boxes, ladders, car seats, sleeping bags, and bottles.
The point cloud captures the three-dimensional structures of objects at varying distances.
Figure~\ref{fig:data_collection_1} shows the point clouds of a chair, an indoor garbage can and a ladder, each color-coded according to height to highlight their spatial dimensions.

\subsection{Postprocessing}

After data collection, the raw information is post-processed to estimate vehicle poses and to anonymize the image data.
Point cloud registration was performed using KISS ICP \cite{vizzo2023kissicp}, a lidar odometry pipeline.
It includes point cloud motion compensation, subsampling, adaptive thresholding to determine correspondences, and lidar pose estimation.
The calculated LiDAR pose was then exported in SemanticKITTI format as required for the labeling tool \cite{behley2019semantickitti}.

\begin{figure}[h!]
    \centering
    \begin{subfigure}[]{\columnwidth}
    \centering
    \includegraphics[trim={0cm 0cm 0cm 0cm},clip,width=\columnwidth]{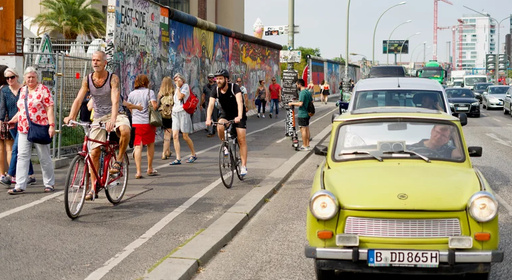}
    \caption{\small Original image (source: \url{https://tinyurl.com/3s66za36})}.
    \label{fig:original_web}
    \end{subfigure}
    \begin{subfigure}[]{\columnwidth}
    \centering
    \includegraphics[trim={0cm 0cm 0cm 0cm},clip,width=\columnwidth]{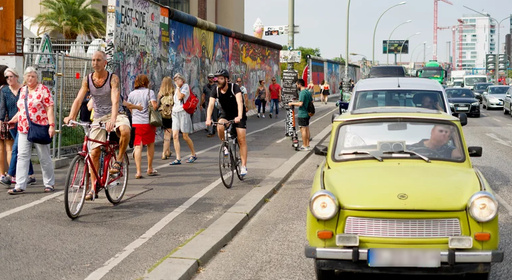}
    \caption{Anonymized image.}
    \label{fig:anonymised_web}
    \end{subfigure}
    \caption{\small Anonymization of camera images.}
    \label{fig:anonymisation}
\end{figure}

Ethical considerations for this dataset include anonymization of camera images to protect individual privacy.
Identifiable information, such as faces and license plates, is processed using DashcamCleaner~\cite{dashcamcleaner} and DeepPrivacy2~\cite{hukkelas23DP2}.
DashcamCleaner uses a license plate detector to locate and blur license plates, while DeepPrivacy2 identifies facial features and generates new unidentifiable photorealistic faces to replace the original ones.
Figure \ref{fig:anonymisation} illustrates the anonymization process using a publicly available image from the internet.

\subsection{Ground Plane Segmentation}

One of the popular approaches for anomaly detection in the point-cloud domain involves applying ground-plane removal algorithms to reduce the search space.
We used Patchwork++~\cite{patchwork} to remove the ground plane from the point-cloud data.
While ground plane removal is effective at short ranges, its performance is reduced at longer distances and on roads with varying geometries.
Under these conditions, Patchwork++ often results in many false positives or incorrectly segments objects as part of the ground plane.
In addition, manually fine-tuning the parameters of such methods to adapt to different topographies is a very challenging task.

\begin{figure}[h!]
    \centering
    \begin{tikzpicture}[spy using outlines={red, magnification=2.5, height=1.5cm, width=2cm, connect spies}]
        \node {\includegraphics[trim={0cm 0cm 0cm 0cm},clip,width=\linewidth]{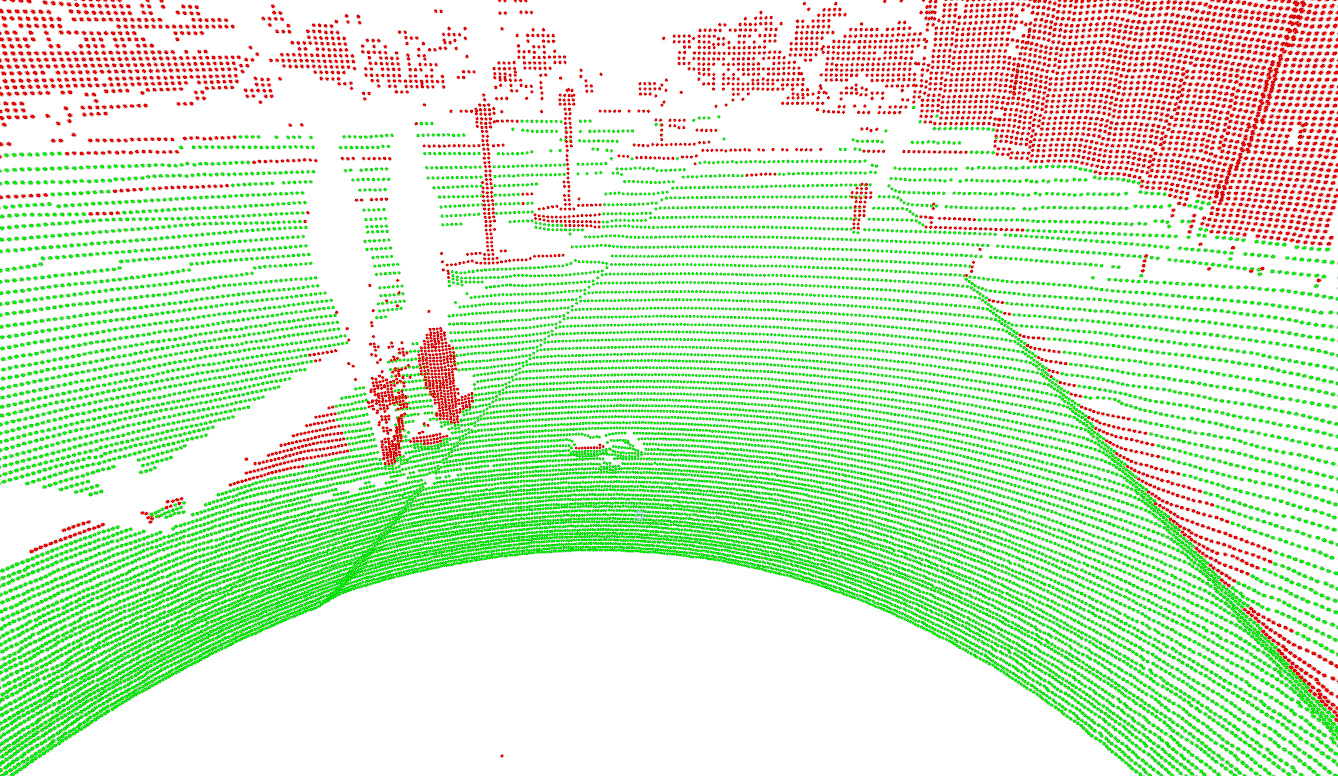}};
        \spy on (-0.4,-0.45) in node [right] at (0.6,0);
    \end{tikzpicture}
    \caption{\small Patchwork++ performance in a wide environment.}
    \label{fig:projection_3}
\end{figure}

\begin{figure}[h!]
    \centering
        \begin{tikzpicture}[spy using outlines={red, magnification=2.5, height=1.5cm, width=2cm, connect spies}]
        \node {\includegraphics[trim={0cm 0cm 0cm 0cm},clip,width=\linewidth]{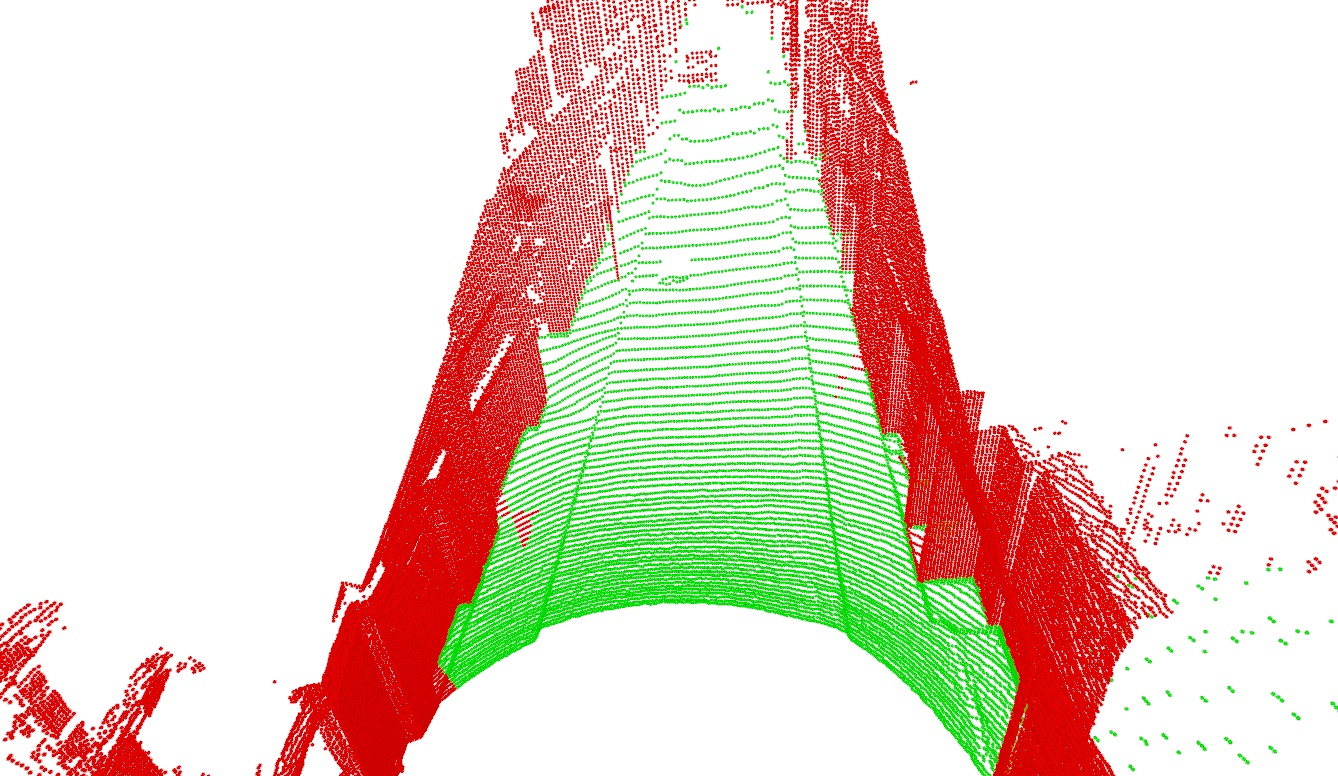}};
        \spy on (-0.0,+0.56) in node [right] at (1.7,0);
    \end{tikzpicture}
    \caption{\small Patchwork++ performance in a narrow urban street.}
    \label{fig:projection_2}
\end{figure}

\section{Low Performance of the 3D Models}

\subsection{Relation of Performance to Distance and Size}
We calculated the AP metric for different distance thresholds, as shown in Table~\ref{tab:distance_ablation}.
In the lower ranges, from $0$ to $10$, and from $10$ to $20$ range models perform better, then at other distances.
Note, that methods evaluate on points within this range, treating points outside of the range as unlabeled, \ie evaluation on $10$--$20$ meters means that points at least as far as $10$ meters are considered for evaluation.
Expectedly, we see a decrease in performance as the distance to the anomaly increases.
In addition, we looked at the relation between the size of the object and anomaly segmentation performance in a sequence.
We observe a drop in performance for small objects in Figure~\ref{fig:objsize}.

\subsection{Number of Foreground Points}
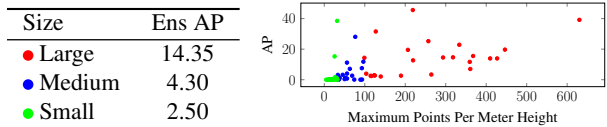
\begin{figure}[t]
    \centering
    \begin{minipage}[b]{0.45\linewidth}
        \centering
        {\small
        \begin{tabular}{lc}
        \toprule
            Size & Ens AP \\
        \midrule
            \colorcirc{red} Large   & 14.35 \\
            \colorcirc{blue} Medium  & 4.30 \\
            \colorcirc{green} Small   & 2.50 \\
        \bottomrule
        \end{tabular}
        }
        \subcaption{Mean AP for the Category.}
        \label{tab:objsize}
    \end{minipage}
    \hfill
    \begin{minipage}[b]{0.5\linewidth}
        \centering
        \resizebox{\linewidth}{!}{
        \LARGE
        \input{figures/8_rebuttal/scatter.tex}
        }
        \subcaption{Mean AP Per Sequence.}
        \label{fig:objsize}
    \end{minipage}
    \caption{Deep Ensembles AP for differently sized objects over validation and test datasets.}
\end{figure}

The class imbalance remains a challenge for Point-Level evaluation, as it is more pronounced in terms of the occupied space and number of points.
If we compare to a SegmentMeIfYouCan setup (see Table 1 from~\cite{chan2021segmentmeifyoucan}), our dataset has $0.03\%$ of anomaly and $36.9\%$ inlier points; that is twice as few anomaly points.
In addition, we evaluate objects with at least 5 anomaly points (instead of 50 or more~\cite{chan2021segmentmeifyoucan}).
However, for some sequences, we observe performance similar to 2D methods, especially for large objects.
On Figure~\ref{tab:objsize}  we show performance of the Ensemble method on a combined validation and test dataset for better illustration.
Here, we split our data into sequences base on the effective size of an object.
We divide the maximum number of points for an instance by the maximum height of an instance in a sequence, and separate sequences into three categories: sequence with an object that has $0-33$ points per meter, $33-99$ and $999+$ points per meter.
We observe that deep ensembles perform better on sequences with larger objects.

\begin{table}[t]
    \centering
    \resizebox{\linewidth}{!}{
    \begin{tabular}{lccccc}
    \toprule
         Method & 0--10m & 10--20m & 20--30m & 30--40m & 40--50m \\
     \midrule
         Deep Ensemble~\cite{lakshminarayanan2017deepensemble}  & 7.63 & 8.49 & 3.42 & 0.38 & 0.03 \\
         MC Dropout~\cite{srivastava2014mcdropout}              & 0.16 & 0.53 & 0.06 & 0.04 & 0.01 \\
         Max Logit~\cite{hendrycks2018baseline}                 & 2.25 & 1.53 & 1.20 & 0.27 & 0.01 \\
         Void Classifier~\cite{blum2021fishyscapes}             & 2.95 & 1.78 & 1.98 & 0.28 & 0.03 \\
         RbA~\cite{nayal2023rba}                                & 1.85 & 1.28 & 0.73 & 0.15 & 0.01 \\
    \bottomrule
    \end{tabular}}
    \caption{
    Anomaly segmentation performance per distance measured by AP.
    }
    \label{tab:distance_ablation}
\end{table}

\begin{table}[t]
    \centering
    {
    \footnotesize
    \begin{tabular}{lcccc}
    \toprule
         Method & Aux Data & AUROC$\uparrow$ & FPR$@95\downarrow$ & AP$\uparrow$ \\
    \midrule
         DenseHybrid~\cite{grcic2022densehybrid} & \xmark & 87.09 & 76.36 & 26.63 \\
         RbA~\cite{nayal2023rba} & \xmark & 89.58 & 75.21 & 37.12 \\ 
         UNO~\cite{delic2024uno} & \cmark & 89.52 & 62.29 & 37.10 \\
         Mask2Anomaly~\cite{rai2023mask2anomaly} & \cmark & 90.54 & 78.09 & 36.38 \\
    \bottomrule
    \end{tabular}
    }
    \caption{Evaluation of 2D methods on our the validation set using only a front-view camera.}
    \label{tab:2deval}
\end{table}

\section{Results on Validation Datasets}

We show results for the SemanticKITTI~\cite{behley2019semantickitti} validation set in Table~\ref{tab:supp-results-semkitti} and our dataset in Table~\ref{tab:supp-results-inlier}.
For the OOD validation set, we evaluate in three sequences and provide scores in Table~\ref{tab:val_set_performance}.

\begin{table*}
\setlength\tabcolsep{3.7pt}
    \begin{center}
    \caption{Class-wise PQ scores on SemanticKITTI validation set.}
        \resizebox{\textwidth}{!}{
        \label{tab:supp-results-semkitti}
        \footnotesize
        \begin{tabular}{l|c|ccccccccccccccccccc|c}
            \toprule
            Method &
            \begin{sideways}void\end{sideways} &
            \begin{sideways}car\end{sideways} &
            \begin{sideways}truck\end{sideways} &
            \begin{sideways}bicycle\end{sideways} &
            \begin{sideways}motorcycle\end{sideways} &
            \begin{sideways}other vehicle\end{sideways} &
            \begin{sideways}person\end{sideways} &
            \begin{sideways}bicyclist\end{sideways} &
            \begin{sideways}motorcyclist\end{sideways} &
            \begin{sideways}road\end{sideways} &
            \begin{sideways}sidewalk\end{sideways} &
            \begin{sideways}parking\end{sideways} &
            \begin{sideways}other ground\end{sideways} &
            \begin{sideways}building\end{sideways} &
            \begin{sideways}vegetation\end{sideways} &
            \begin{sideways}trunk\end{sideways} &
            \begin{sideways}terrain\end{sideways} &
            \begin{sideways}fence\end{sideways} &
            \begin{sideways}pole\end{sideways} &
            \begin{sideways}traffic sign\end{sideways} &
            PQ \\
            \midrule
            Mask-PLS$^*$~\cite{marcuzzi2023maskpls}             & -- & 94.12 & 83.19 & 44.55 & 61.44 & 59.36 & 77.08 & 91.64 & 0.0 & 93.94 & 78.67 & 37.19 & 0.0 & 86.99 & 84.85 & 53.28 & 57.06 & 21.35 & 59.85 & 53.62 & 59.9 \\
            Mask-PLS~\cite{marcuzzi2023maskpls}             & -- & 91.90 & 77.93 & 16.96& 51.28 & 45.66 & 65.85 & 83.36 & 0.0 & 93.86 & 77.69 & 31.39 & 0.0 & 86.97 & 87.61 & 50.40 & 59.10 & 22.79 & 60.77 & 53.34 & 55.62 \\
            Mask4Former-3D$^*$~\cite{yilmaz2024mask4former} & -- & 93.81 & 70.95 & 62.92 & 68.97 & 56.79 & 81.98 & 87.35 & 24.06 & 93.94 & 78.05 & 27.33 & 0.0 & 88.39 & 88.65 & 50.93 & 60.82 & 25.38 & 57.89 & 58.59 & 61.94\\
            Mask4Former-3D                                  & -- & 93.53 & 59.39 & 62.55 & 64.82 & 54.36 & 79.61 & 89.16 & 25.01 & 93.24 & 77.90 & 28.79 & 0.0 & 87.27 & 87.28 & 51.08 & 59.92 & 24.85 & 56.76 & 58.14 & 60.72\\
            Mask4Former-3D-void                             & 6.08 & 74.36 & 47.00 & 32.19 & 43.34 & 33.30 & 42.90 & 68.75 & 00.33 & 93.35 & 77.07 & 19.01 & 0.0 & 82.77 & 81.34 & 47.56 & 56.94 & 19.98 & 54.48 & 36.82 & 47.97 \\
            \bottomrule
        \end{tabular}
        }
    \end{center}
\end{table*}

\begin{table*}
\setlength\tabcolsep{3.7pt}
    \begin{center}
    \caption{Class-wise PQ scores on STU-inlier Validation Set.}
        \resizebox{\textwidth}{!}{
        \label{tab:supp-results-inlier}
        \footnotesize
        \begin{tabular}{l|c|cccccccccccccc|c}
            \toprule
            Method &
            \begin{sideways}void\end{sideways} &
            \begin{sideways}car\end{sideways} &
            \begin{sideways}truck\end{sideways} &
            \begin{sideways}bicycle\end{sideways} &
            \begin{sideways}person\end{sideways} &
            \begin{sideways}road\end{sideways} &
            \begin{sideways}sidewalk\end{sideways} &
            \begin{sideways}parking\end{sideways} &
            \begin{sideways}building\end{sideways} &
            \begin{sideways}vegetation\end{sideways} &
            \begin{sideways}trunk\end{sideways} &
            \begin{sideways}terrain\end{sideways} &
            \begin{sideways}fence\end{sideways} &
            \begin{sideways}pole\end{sideways} &
            \begin{sideways}traffic sign\end{sideways} &
            PQ \\
            \midrule
            Mask-PLS$^*$~\cite{marcuzzi2023maskpls} & -- &78.88 & 1.89 & 0.0 & 70.92 & 56.34 & 26.18 & 0.0 & 54.96 & 74.69 & 1.36 & 55.20 & 48.00 & 41.51 & 44.51 &39.60\\
            Mask-PLS & -- & 78.66 & 22.16 & 0.0 & 75.33 & 81.77 & 41.46& 0.0 & 78.79 & 89.16 & 49.92 & 25.53 & 46.53 &56.79 & 66.94& 50.93\\
            Mask4Former-3D$^*$~\cite{yilmaz2024mask4former} & -- & 78.45 &11.29 &10.59 & 69.44 & 59.07 & 41.99 & 0.27 & 84.70 & 88.46 & 0.0 & 0.0 & 36.51 & 25.92 & 57.50 & 42.80\\
            Mask4Former-3D & -- & 80.99 & 37.28 & 47.65 & 80.99 & 71.46 & 17.74 & 0.0 & 84.08 & 89.73 & 29.34 & 30.79 & 47.6 & 59.62 & 60.96 & 52.73 \\
            Mask4Former-3D-void & 0.07 & 23.88 & 20.78 & 1.01 & 43.30 & 38.24 & 20.03 & 11.11 & 48.45 & 43.09 & 20.20 & 17.31 & 30.80 & 27.26 & 33.16 & 26.96 \\
            \bottomrule
        \end{tabular}
        }
    \end{center}
\end{table*}

\begin{table*}[t]
    \centering
    \caption{\small Anomaly Segmentation Performance on the Validation Set with Anomalies}
    \resizebox{\textwidth}{!}{
    \begin{tabular}{lcccccccccc}
        \toprule
        \multirow{2}{*}{Method} & \multicolumn{3}{c}{Point-Level OOD} && \multicolumn{5}{c}{Object-Level OOD} \\
        \cline{2-4}
        \cline{6-10}
          & AUROC~$\uparrow$ & FPR@95~$\downarrow$ & AP~$\uparrow$ && RecallQ & SQ & RQ & UQ & PQ \\
        \midrule
        Deep Ensemble~\cite{lakshminarayanan2017deepensemble}  & 90.93 & 37.34 & 6.94 && 17.70 & 79.96 & 9.10 & 14.15 & 7.27 \\
        MC Dropout~\cite{srivastava2014mcdropout}           & 65.76 & 79.82 & 0.17 && 3.54 & 74.36 & 3.48 & 2.63 & 2.59 \\
        Max Logit~\cite{hendrycks2018baseline}           & 87.27 & 68.76 & 2.02 && 26.64 & 79.26 & 2.06 & 21.12 & 1.63 \\
        Void Classifier~\cite{blum2021fishyscapes}  & 89.77 & 79.50 & 2.62 && 17.35 & 81.27 & 8.98 & 14.10 & 7.30 \\
        RbA~\cite{nayal2023rba}                      & 73.00 & 100.0 & 1.64 && 21.84 & 78.58 & 2.75 & 17.16 & 2.16 \\
        \bottomrule
    \end{tabular}}
    \label{tab:val_set_performance}
\end{table*}

\subsection{Evaluation of 2D Methods}
We focus on automotive applications, where it is common practice to evaluate multimodal methods on 3D LiDAR annotations, since 3D distances to objects are important for driving.
As a control experiment, we applied 2D-only methods to the frontal camera and evaluated only within the corresponding frustum of LiDAR points (see Table~\ref{tab:2deval}).
Methods that use only RGB images have a higher false-positive rate in this setup compared to 2D benchmarks.
We attribute this to a domain gap of the Cityscapes and SemanticKitti, partially because of label conventions, \ie, parking lots or backs of traffic signs are ``unlabeled"\ in Cityscapes but are ``inlier"\ in SemanticKitti and these regions contribute to higher FPR.
As well as to the LiDAR-Camera points misalignment, \ie at image boundaries.
However, in our dataset, anomalies appear in images from cameras with other perspectives, and evaluating a full 360-degree view would be more difficult.

\section{Annotation and Qualitative Examples}
\begin{figure*}
    \centering
    \includegraphics[width=\textwidth]{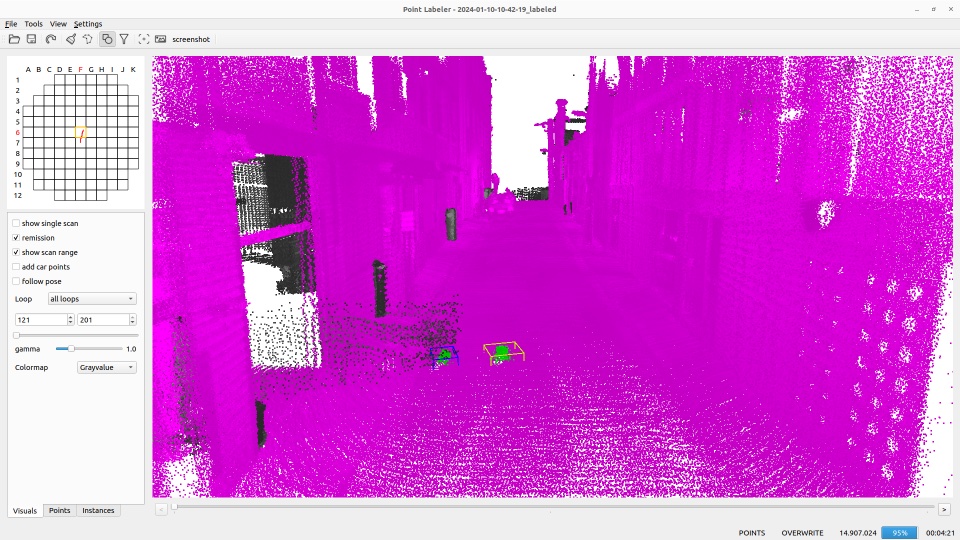}
    \caption{
    Data Annotation Example: Each color represents a specific label — Purple for inlier, Green for anomaly, and Black for void.
    Boxes represent instance boundaries.
    }
    \label{fig:annotation-example}
\end{figure*}

\begin{figure*}[ht]
    \centering
    \begin{subfigure}[t]{0.24\textwidth}
        \includegraphics[width=\linewidth]{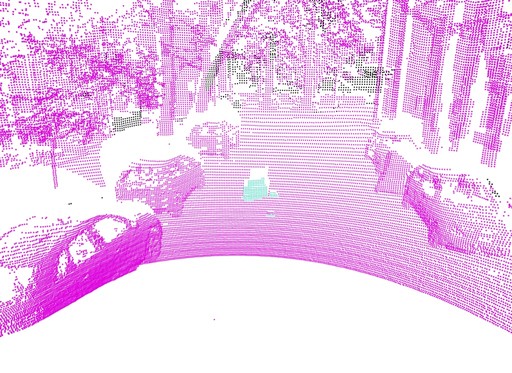}
    \end{subfigure}
    \begin{subfigure}[t]{0.24\textwidth}
        \includegraphics[width=\linewidth]{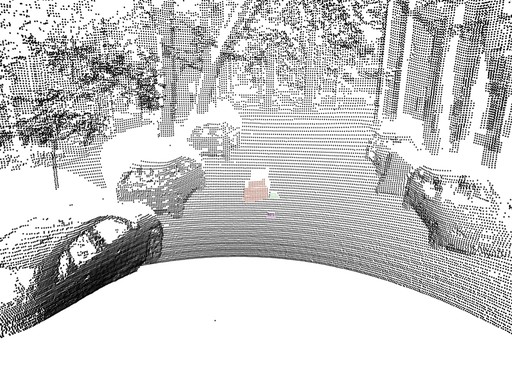}
    \end{subfigure}
    \begin{subfigure}[t]{0.24\textwidth}
        \includegraphics[width=\linewidth]{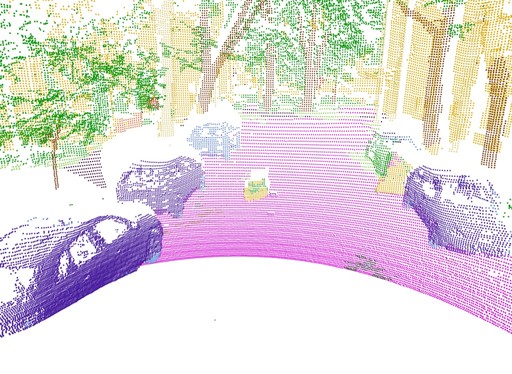}
    \end{subfigure}
    \begin{subfigure}[t]{0.24\textwidth}
        \includegraphics[width=\linewidth]{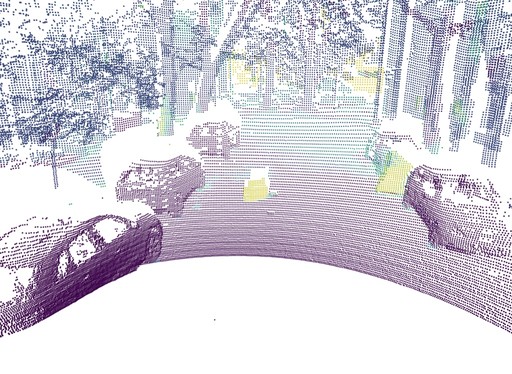}
    \end{subfigure}

    \begin{subfigure}[t]{0.24\textwidth}
        \includegraphics[width=\linewidth]{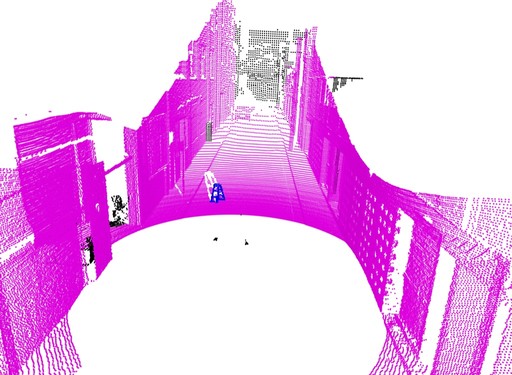}
    \end{subfigure}
    \begin{subfigure}[t]{0.24\textwidth}
        \includegraphics[width=\linewidth]{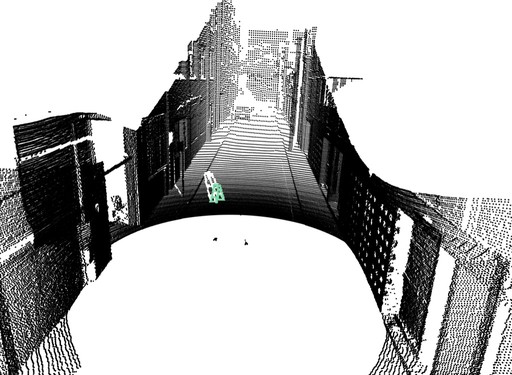}
    \end{subfigure}
    \begin{subfigure}[t]{0.24\textwidth}
        \includegraphics[width=\linewidth]{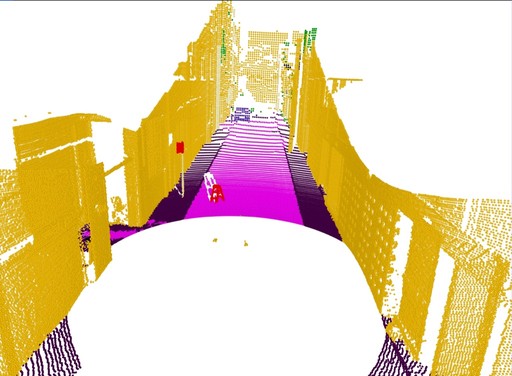}
    \end{subfigure}
    \begin{subfigure}[t]{0.24\textwidth}
        \includegraphics[width=\linewidth]{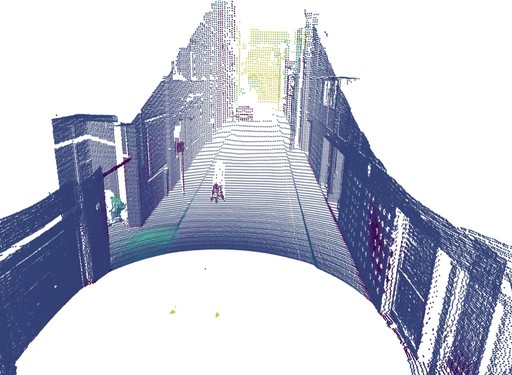}
    \end{subfigure}

    \begin{subfigure}[t]{0.24\textwidth}
        \includegraphics[width=\linewidth]{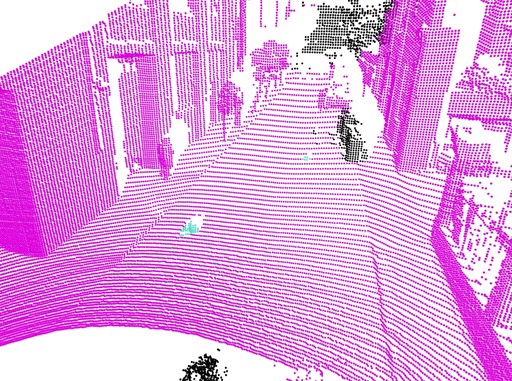}
    \end{subfigure}
    \begin{subfigure}[t]{0.24\textwidth}
        \includegraphics[width=\linewidth]{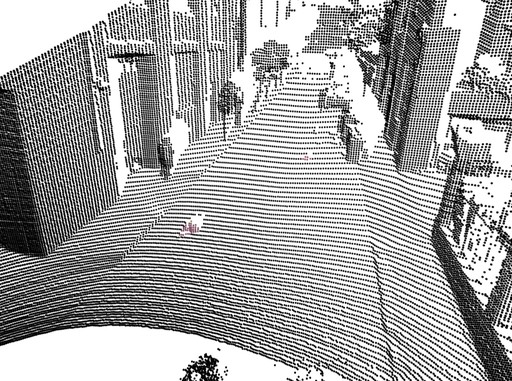}
    \end{subfigure}
    \begin{subfigure}[t]{0.24\textwidth}
        \includegraphics[width=\linewidth]{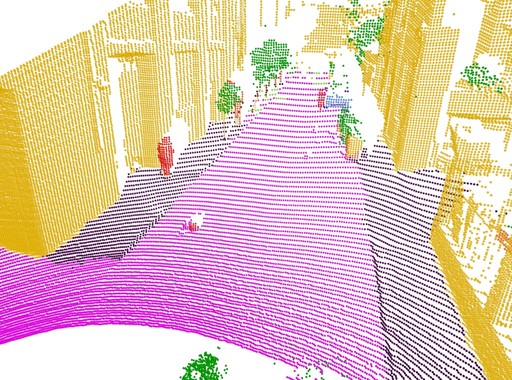}
    \end{subfigure}
    \begin{subfigure}[t]{0.24\textwidth}
        \includegraphics[width=\linewidth]{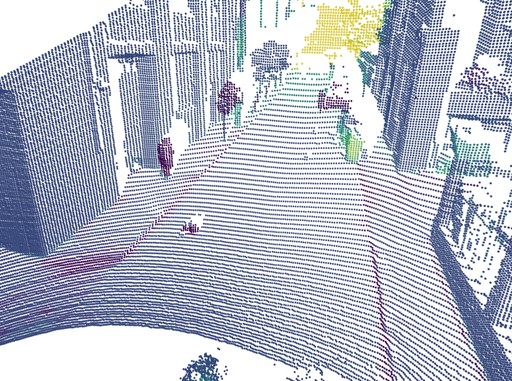}
    \end{subfigure}

    \begin{subfigure}[t]{0.24\textwidth}
        \includegraphics[width=\linewidth]{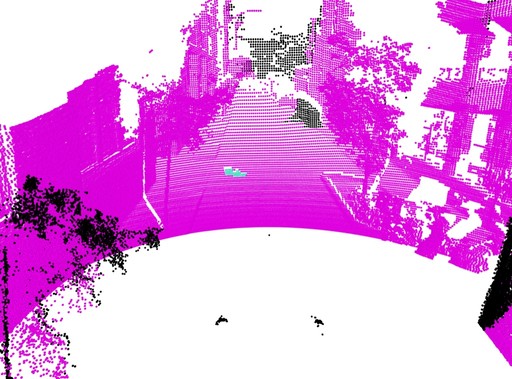}
    \end{subfigure}
    \begin{subfigure}[t]{0.24\textwidth}
        \includegraphics[width=\linewidth]{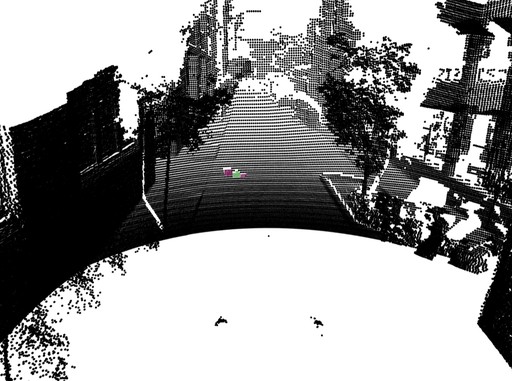}
    \end{subfigure}
    \begin{subfigure}[t]{0.24\textwidth}
        \includegraphics[width=\linewidth]{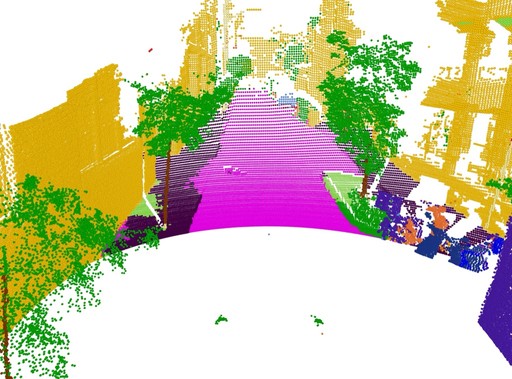}
    \end{subfigure}
    \begin{subfigure}[t]{0.24\textwidth}
        \includegraphics[width=\linewidth]{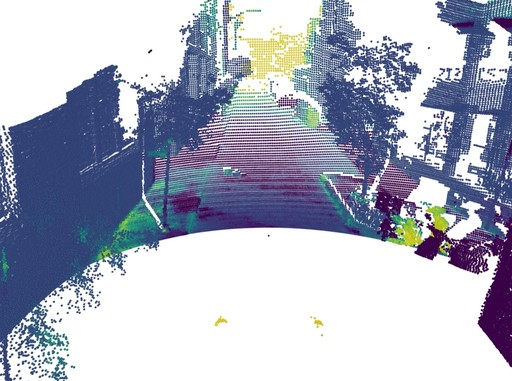}
    \end{subfigure}

    \begin{subfigure}[t]{0.24\textwidth}
        \includegraphics[width=\linewidth]{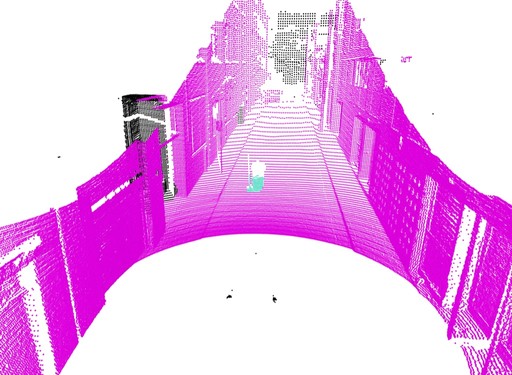}
    \end{subfigure}
    \begin{subfigure}[t]{0.24\textwidth}
        \includegraphics[width=\linewidth]{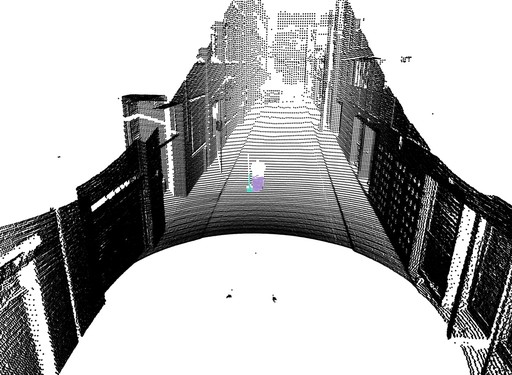}
    \end{subfigure}
    \begin{subfigure}[t]{0.24\textwidth}
        \includegraphics[width=\linewidth]{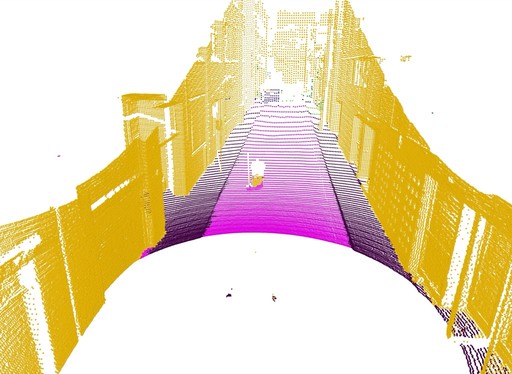}
    \end{subfigure}
    \begin{subfigure}[t]{0.24\textwidth}
        \includegraphics[width=\linewidth]{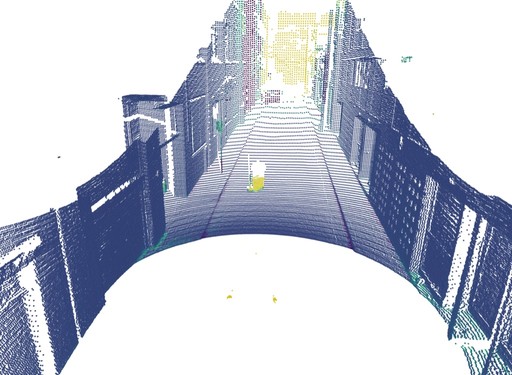}
    \end{subfigure}

    \begin{subfigure}[t]{0.24\textwidth}
        \includegraphics[width=\linewidth]{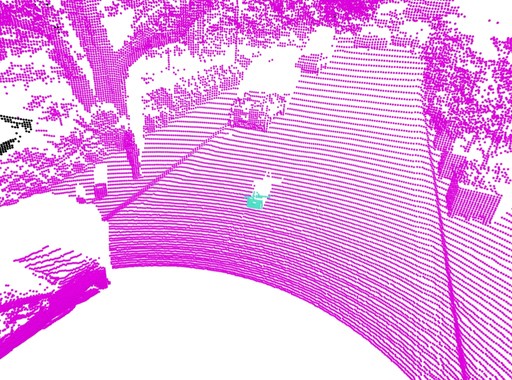}
    \end{subfigure}
    \begin{subfigure}[t]{0.24\textwidth}
        \includegraphics[width=\linewidth]{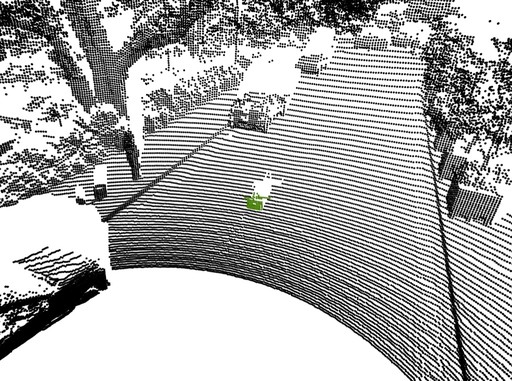}
    \end{subfigure}
    \begin{subfigure}[t]{0.24\textwidth}
        \includegraphics[width=\linewidth]{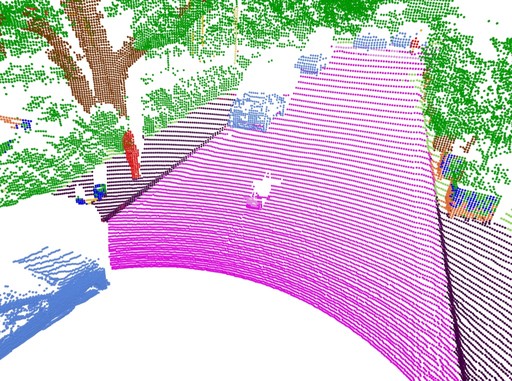}
    \end{subfigure}
    \begin{subfigure}[t]{0.24\textwidth}
        \includegraphics[width=\linewidth]{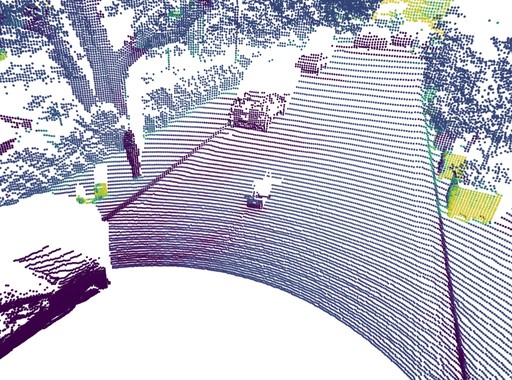}
    \end{subfigure}

    \begin{subfigure}[t]{0.24\textwidth}
        \includegraphics[width=\linewidth]{figures/7_supplementary/qualitative_results/0_label.jpg}
        \caption{Anomaly Label.}
    \end{subfigure}
    \begin{subfigure}[t]{0.24\textwidth}
        \includegraphics[width=\linewidth]{figures/7_supplementary/qualitative_results/0_label_instance.jpg}
        \caption{Instance Label.}
    \end{subfigure}
    \begin{subfigure}[t]{0.24\textwidth}
        \includegraphics[width=\linewidth]{figures/7_supplementary/qualitative_results/0_prediction.jpg}
        \caption{Inlier Prediction.}
    \end{subfigure}
    \begin{subfigure}[t]{0.24\textwidth}
        \includegraphics[width=\linewidth]{figures/7_supplementary/qualitative_results/0_rba.jpg}
        \caption{RBA.}
    \end{subfigure}

    \caption{
    Visualization of the proposed dataset with anomaly labels, instance labels, inlier class predictions, and anomaly scores of the selected anomaly methods.
    }
    \label{fig:supp_qualitative_results}
\end{figure*}

We visualize the annotation interface with an example of a correctly annotated scene in the figure~\ref{fig:annotation-example}.
Anomaly points are cyan, unlabeled regions are black, and inliers are purple.
We provide further visualizations of the dataset and the predictions shown in the Figure~\ref{fig:supp_qualitative_results}.

\section{SemanticKITTI Other-object Examples}
\begin{figure*}
    \centering
    \includegraphics[width=0.6\textwidth]{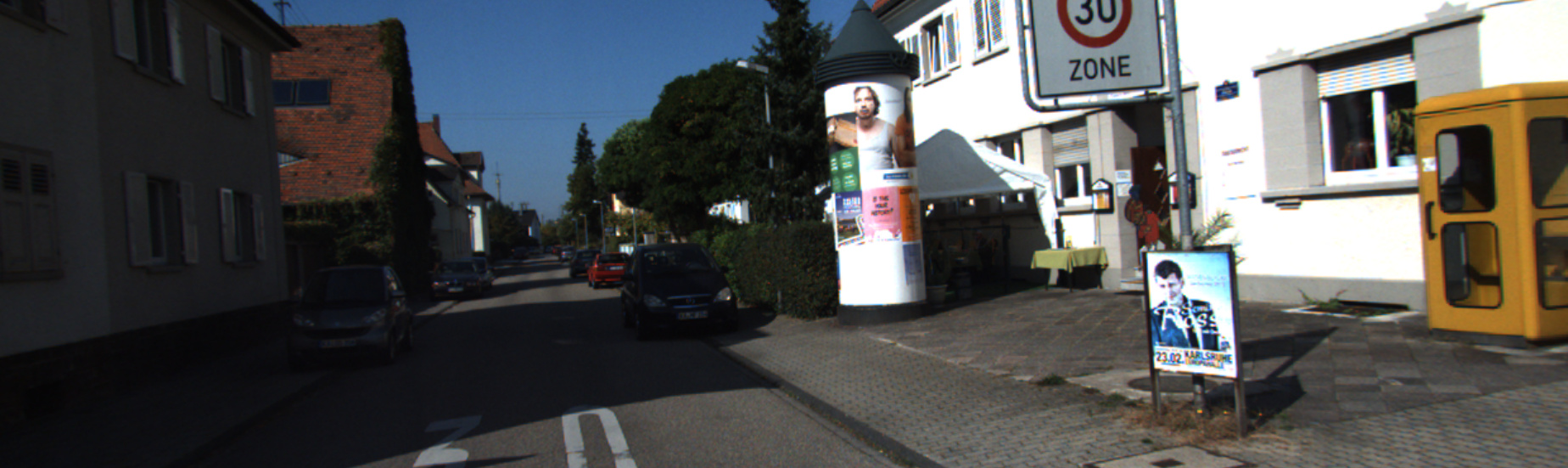}
    \includegraphics[width=0.6\textwidth]{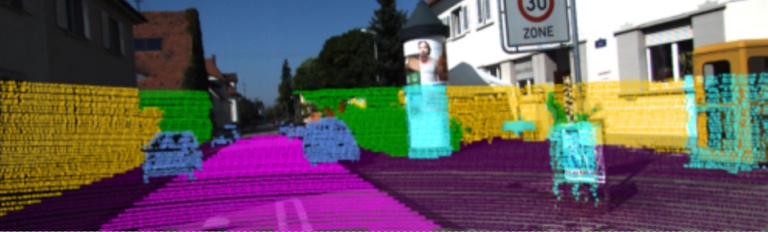}

    \caption{
    Example of the other-object class: a billboard, a smaller billboard, a phone booth, and a small table, all of which belong to the other-object class.
    }
    \label{fig:other-obj}
\end{figure*}

\begin{figure*}
    \centering
    \includegraphics[width=0.6\textwidth]{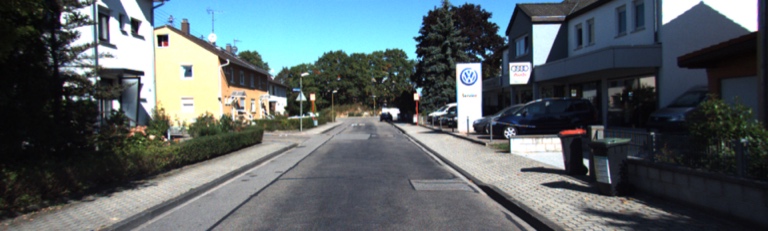}
    \includegraphics[width=0.6\textwidth]{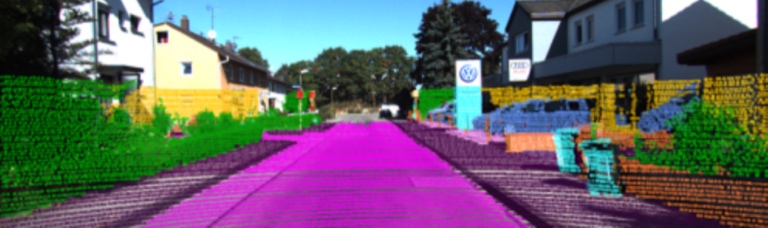}

    \caption{
    Example of the other-object class: a car dealership sign and two garbage cans, all belonging to the other-object class.
    }
    \label{fig:other-obj2}
\end{figure*}

\begin{figure*}
    \centering
    \includegraphics[width=0.6\textwidth]{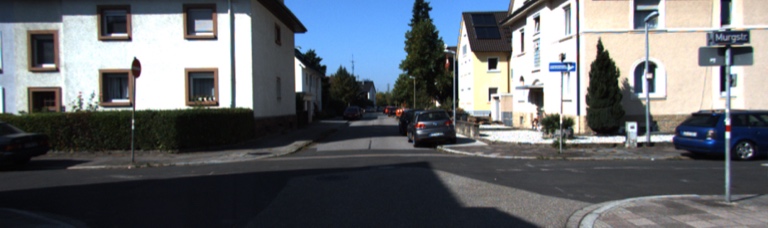}
    \includegraphics[width=0.6\textwidth]{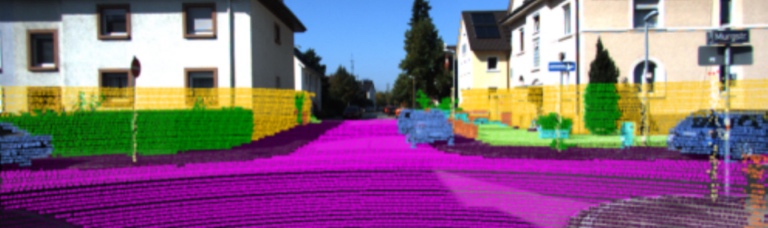}
    \caption{
    Example of the other-object class: a potted plant and a power adapter, all of which belong to the other-object class.
    }
    \label{fig:other-obj3}
\end{figure*}

Several examples of the other-object class in the SemanticKITTI dataset can be seen in Figure~\ref{fig:other-obj}, Figure~\ref{fig:other-obj2}, and Figure~\ref{fig:other-obj3}.
The other-object class consists of many miscellaneous items, including trash bins, advertisement posts, and small pots.
We superimposed the class label on the front-facing camera, with light blue denoting the other object class.

\section{Note on Training}
Initially, jointly training with both SemanticKITTI and Panoptic-CUDAL led to diverging losses for Mask4Former-3D.  This also occurred during training runs solely on Panoptic-CUDAL. Lowering the preset learning rate from 0.0004 to 0.0002 was enough to mitigate the loss divergence in both cases.

%% file: figures/8_rebuttal/scatter.tex
\begin{tikzpicture}[trim axis left, trim axis right, clip]
    \begin{axis}[
        width=14cm,
        height=5cm,
        xlabel={Maximum Points Per Meter Height},
        ylabel={AP},
    ]

    \addplot[
        only marks,
        mark=*,
        color=red
    ] coordinates {
    (123.20810770988464,2.915209225845219)
    (127.17485904693604,31.551439093492334)
    (189.54638016223907,2.636370843873389)
    (293.4390985965729,14.541066079793735)
    (116.38233661651611,2.405329025458638)
    (630.2624621391296,39.17266625248943)
    (218.70064508914948,45.581867979054216)
    (333.4930624961853,22.864470917129996)
    (114.46899175643921,2.551745351938195)
    (360.647212266922,7.02654932004233)
    (139.4078038930893,2.102556843595984)
    (264.5034942626953,3.3938312725411994)
    (219.35696125030518,12.668406652395984)
    (317.9991645812988,14.64510737899587)
    (427.00393295288086,13.915823278576179)
    (367.7030372619629,15.521905472680661)
    (257.07945799827576,25.210041257331827)
    (359.3410634994507,11.672730362924462)
    (124.89179992675781,2.732880089011521)
    (409.69018363952637,13.919924482023161)
    (103.61346817016602,3.9370488120788845)
    (446.335355758667,19.692394786905233)
    (206.5576148033142,19.53222805191654)
    (99.05119585990906,14.342928890196987)
    };

    \addplot[
        only marks,
        mark=*,
        color=blue
    ] coordinates {
    (89.25047588348389,0.03098103887641902)
    (63.07181739807129,7.106650917581521)
    (56.00456714630127,11.186514286488437)
    (45.50699281692505,3.2007005004087152)
    (56.154481172561646,4.027849781001339)
    (35.68193006515503,0.5489912488926904)
    (96.56789946556091,11.796577430624154)
    (51.96006774902344,1.3030696773771355)
    (36.10545003414154,0.1413604466743647)
    (33.51892614364624,3.1367917994459114)
    (36.96332931518555,1.5788870290746306)
    (50.14889395236969,0.4074992174530348)
    (93.26884031295776,0.10090122818761124)
    (68.8652515411377,2.634610220633483)
    (37.922985315322876,1.0439469052423245)
    (56.223361015319824,0.7718033113881734)
    (34.2248729467392,1.3718722818675877)
    (74.06554841995239,0.07139276863832736)
    (76.52416229248047,28.067624873817167)
    (92.79764604568481,7.53855637045290)
    };

    \addplot[
        only marks,
        mark=*,
        color=green
    ] coordinates {
    (16.28862953186035,0.09345251383246322)
    (9.696988105773926,0.10760287553699889)
    (31.683093309402466,38.48258234698749)
    (30.575931787490845,1.1650389553740177)
    (13.966336011886597,0.33556633507454536)
    (13.708930850028992,0.0731694535156185)
    (20.034308433532715,0.07046463511257164)
    (17.79423236846924,0.034373296555552504)
    (16.44211721420288,0.1663483939183421)
    (28.911840677261353,0.7921602525646146)
    (18.535447597503662,0.019018406570544214)
    (25.44390296936035,15.253311055376633)
    (26.41380214691162,1.6324877100543547)
    (21.68398427963257,0.04388474182709722)
    (8.374138355255127,0.018693220178917527)
    (32.29384660720825,0.14337858140449172)
    (29.06564712524414,0.06609864676908923)
    (13.279111385345459,0.3260171515527739)
    (4.631475448608398,0.01631369112497827)
    (15.716129899024963,0.015446377717895)
    (16.24239444732666,0.05890233369832644)
    (25.06314468383789,0.1933458263095459)
    (31.887524604797363,0.02471219334070912)
    (25.914663076400757,0.38242440423472585)
    (15.145784139633179,0.20436300069802543)
    };

    \end{axis}
\end{tikzpicture}